\documentclass{article}

\usepackage{graphicx,amscd,amsmath,amssymb,verbatim}
\usepackage[dvips]{hyperref}
\usepackage[TS1,OT1,T1]{fontenc}

\begin{document}

\title{Discrete Network Dynamics. Part 1: Operator Theory}
\author{Stephen Luttrell}
\maketitle
\begin{abstract}

An operator algebra implementation of Markov chain Monte Carlo algorithms
for simulating Markov random fields is proposed. It allows the dynamics
of networks whose nodes have discrete state spaces to be specified
by the action of an update operator that is composed of creation
and annihilation operators. This formulation of discrete network
dynamics has properties that are similar to those of a quantum field
theory of bosons, which allows reuse of many conceptual and theoretical
structures from QFT. The equilibrium behaviour of one of these generalised
MRFs and of the adaptive cluster expansion network (ACEnet) are
shown to be equivalent, which provides a way of unifying these two
theories.
\end{abstract}
\section{Introduction}\label{XRef-Section-116135347}

The aim of this paper is to present a theoretical framework for
building recurrent network models where the states of the network
nodes are discrete-valued, which will define a general framework
for discrete information processing that can be implemented in various
computational architectures. The introduction of recurrence into
networks makes them much more difficult to analyse and control than
feed-forward networks. The basic reason for these difficulties is
that loopy propagation in recurrent networks causes each network
observable to be a sum of an infinite (or, at least, a very large)
number of contributions.

One type of network that can be modelled using this framework is
a network of spiking neurons, where the presence or absence of a
spike is a binary quantity (i.e. it is discrete-valued). However,
in this paper, there is no specific aim to model biological information
processing, but there will nevertheless be points of contact between
the general information processing framework presented here and
the specific details of biological information processing.

The only consistent way of processing information is to use Bayesian
methods \cite{Cox1946}, which represent information by using\ \ the
joint probability of the states of the network nodes, and process
information (or make inferences) by manipulating these joint probabilities
according to well-defined rules such as Bayes theorem. The Bayesian
approach achieves its consistency by \textit{not} discarding any
of the various alternative inferences that can be made, and by following
up the consequences of all of the alternatives it ensures that there
are never any of the contradictions that would otherwise occur,
such as reaching conclusions that depend on which route one takes
through the maze of inferences.

Bayesian information processing needs a flexible way of representing
and manipulating joint probabilities. An ideal framework for this
is Markov random field (MRF) theory \cite{Smyth1998}, because it
allows one to systematically build up a joint probability model
out of pieces that have a simple functional dependence on the underlying
state variables. For networks that have a finite number of nodes,
each of which has a finite number of states, the MRF approach allows
\textit{all} possible joint probability models to be constructed,
so use of the MRF framework imposes no artificial constraints. Because
the MRF approach constructs a joint probability model, it can be
cleanly coupled to any other probability modelling approach.

The implementation of MRFs is usually done using stochastic Markov
chain Monte Carlo (MCMC) computations, unless the MRF happens to
have a particularly simple topology which allows a simpler deterministic
implementation to be achieved (e.g. a tree-like topology allows
exact computations to be done). In this paper no simplifying assumptions
will be made about the network topology, in order to create the
most general possible theoretical framework for discrete information
processing. The simplest type of MCMC computation stochastically
updates the joint state of the MRF, so that it moves around its
joint state space visiting every joint state with a frequency that
is proportional to the joint probability specified by the MRF. More
sophisticated MCMC computations do the same thing but with an \textit{ensemble}
of joint states of the MRF; these are known as ``particle filtering''
algorithms \cite{DoucetGodsillAndrieu2000}.

The main result that is presented in this paper is a new way of
describing MCMC algorithms, in which the updating of the MRF joint
state (i.e. the joint state of the network nodes) is decomposed
into a set of more elementary operations, which are the creation
and annihilation of network node states. In the simplest case, a
single MCMC update changes the joint state of an MRF by modifying
its state at a single node of the network, which can be decomposed
into first annihilating the old node state then creating the new
node state. Any MCMC algorithm can be composed out of a sequence
of such creation and annihilation operations. Furthermore, the properties
of the operators that enact these creation and annihilation operations
are very familiar to physicists, because they are identical to the
properties of the creation and annihilation operators that appear
in a quantum field theory (QFT) of bosons \cite{Zee2003}. This allows
a lot of prexisting conceptual and computational machinery to be
brought to bear upon the problem of describing MCMC algorithms.
By drawing an analogy with multi-particle QFT states, the MRF framework
can be consistently generalised so that each node of the network
exists in a \textit{multiply} occupied state, rather than a \textit{singly}
occupied state. There are also many other points of contact with
QFT.

The generalisation of the MRF framework to multiply occupied node
states allows contact to be made with a particular type of self-organising
network (SON) theory known as the adaptive cluster expansion network
(ACEnet) \cite{Luttrell1996}. One of the aims of a SON is to discover
for itself what network architecture to use to solve an information
processing task, so it must be able to dynamically change its architecture.
This requires splitting and merging of network nodes, and also the
creation of appropriate links between them. In an MRF, if a node
is split into two nodes there is no consistent way of assigning
a pairwise state to the resulting pair of nodes, unless the preexisting
single node had two (or more) states assigned to it in the first
place. This is exactly what multiple occupancy in the generalised
MRF framework provides, using creation and annihilation operators
to manipulate these states. Thus the creation and annihilation operator
approach allows MRF theory and SON theory can be cleanly unified.

The structure of this paper is as follows. In Section \ref{XRef-Section-41134323}
the theory of MRFs is summarised, together with the details of MCMC
algorithms for simulating MRFs. In Section \ref{XRef-Subsection-33113632}
the main new contribution of this paper is presented, which is an
operator implementation of the MCMC algorithm that generalises MRF
theory to multiple occupancy states. Finally, in Section \ref{XRef-Section-11423193}
some simple applications are used to illustrate the use of this
operator implementation, one of which is the demonstration that
the equilibrium state of a particular type of multiply occupied
MRF has the same properties as ACEnet.
\section{Markov Random Fields}\label{XRef-Section-41134323}

The aim of this section is to review the MRF framework for building
and manipulating the joint probability models that are used when
doing Bayesian information processing. This includes some informal
material in which multiple occupancy of node states is discussed
before giving the more formal development later on in Section \ref{XRef-Subsection-33113632}.

Section \ref{XRef-Subsubsection-331132456} introduces MRFs and the
Hammersley-Clifford expansion of joint probabilities, and Section
\ref{XRef-Subsubsection-4611524} describes an MCMC algorithm for
sampling the joint states of an MRF. Section \ref{XRef-Subsubsection-331134331}
introduces the concept of a multiple occupancy state which is essential
for the generalisation of MRFs that is presented later in Section
\ref{XRef-Subsection-33113632}. Finally, Section \ref{XRef-Subsubsection-4712833}
describes how MRFs can be used to do Bayesian inference.
\subsection{Basic Markov Random Field Theory}\label{XRef-Subsubsection-331132456}

MRFs are a flexible way of constructing joint probabilities based
on the Hammersley-Clifford expansion (HCE), which is defined as
\cite{Besag1974}
\begin{equation}
\Pr ( \textbf{\textit{x}}) =\frac{1}{Z}\prod _{k}\prod _{c}p_{c}^{k}(
{\textbf{\textit{x}}}_{c}) %
\label{XRef-Equation-322152418}
\end{equation}

\noindent where $\textbf{\textit{x}}$ is the joint state $(x_{1},x_{2},\cdots
,x_{N})$ of an MRF with $N$ nodes, $k$ is the order of the term
in the expansion (i.e. $k$ is the number of components of $\textbf{\textit{x}}$
that the term depends on, which is thus a $k$-tuple), $c$ labels
the particular $k$-tuple (or $k$-clique) that the term depends on,
${\textbf{\textit{x}}}_{c}$ is the $k$-tuple (or clique state),
$p_{c}^{k}( {\textbf{\textit{x}}}_{c}) $ is the probability factor
(or clique factor) associated with ${\textbf{\textit{x}}}_{c}$,
and $Z$ is a normalisation factor to ensure that the total probability
sums to unity as $\sum _{\textbf{\textit{x}}}\Pr ( \textbf{\textit{x}})
=1$, so $Z$ is defined as
\begin{equation}
Z\equiv \sum _{\textbf{\textit{x}}}\prod _{k}\prod _{c}p_{c}^{k}(
{\textbf{\textit{x}}}_{c}) %
\label{XRef-Equation-1018135026}
\end{equation}

\noindent There are some minor technical issues to do with exactly
how the states of the ${\textbf{\textit{x}}}_{c}$ are enumerated
in the HCE to ensure that states are not double-counted, but these
are not important here.

To compute the average $\langle \textbf{\textit{S}}\rangle $ of
a statistic $\textbf{\textit{S}}( \textbf{\textit{x}}) $ you need
to evaluate the following
\begin{equation}
\begin{array}{rl}
 \left\langle  \textbf{\textit{S}}\right\rangle   & =\sum _{\textbf{\textit{x}}}\Pr
( \textbf{\textit{x}}) \textbf{\textit{S}}( \textbf{\textit{x}})
\\
  & =\frac{\sum _{\textbf{\textit{x}}}\prod _{k}\prod _{c}p_{c}^{k}(
{\textbf{\textit{x}}}_{c}) \textbf{\textit{S}}( \textbf{\textit{x}})
}{\sum _{\textbf{\textit{x}}}\prod _{k}\prod _{c}p_{c}^{k}( {\textbf{\textit{x}}}_{c})
}
\end{array}%
\label{XRef-Equation-322151345}
\end{equation}

\noindent where the probability factor $\Pr ( \textbf{\textit{x}})
$ appropriately weights the contribution of each $\textbf{\textit{x}}$
in the sum, so that overall the correct weighted average $\langle
\textbf{\textit{S}}\rangle $ is computed. Despite the functional
simplicity of the HCE expression for $\Pr ( \textbf{\textit{x}})
$, it is usually \textit{not} possible to evaluate Equation \ref{XRef-Equation-322151345}
in closed-form, so numerical techniques must be used.

An intuitive feel for how Equation \ref{XRef-Equation-322151345}
can be evaluated can be obtained by noting that the {\itshape relative}
probability of a pair of joint states ${\textbf{\textit{x}}}_{1}$
and ${\textbf{\textit{x}}}_{2}$ is given by
\begin{equation}
\frac{\Pr ( {\textbf{\textit{x}}}_{1}) }{\Pr ( {\textbf{\textit{x}}}_{2})
}=\frac{\prod _{k}\prod _{c}p_{c}^{k}( {\left( {\textbf{\textit{x}}}_{1}\right)
}_{c}) }{\prod _{k}\prod _{c}p_{c}^{k}( {\left( {\textbf{\textit{x}}}_{2}\right)
}_{c}) }%
\label{XRef-Equation-47123935}
\end{equation}

\noindent where the normalising $Z$ factor in Equation \ref{XRef-Equation-322152418}
cancels, and also any factors in common between the numerator and
denominator of the ratio in Equation \ref{XRef-Equation-47123935}\ \ will
cancel. Thus, if the joint states ${\textbf{\textit{x}}}_{1}$ and
${\textbf{\textit{x}}}_{2}$ differ in only a few of their vector
components, then any of the probability factors $p_{c}^{k}( {\textbf{\textit{x}}}_{c})
$ that do \textit{not} depend on these differing components will
cancel out, leaving a relatively simple expression for the ratio
$\frac{\Pr ( {\textbf{\textit{x}}}_{1}) }{\Pr ( {\textbf{\textit{x}}}_{2})
}$. This cancellation is a key property of the functional form of
the HCE in Equation \ref{XRef-Equation-322152418}. Once a simple
expression for the relative probability $\frac{\Pr ( {\textbf{\textit{x}}}_{1})
}{\Pr ( {\textbf{\textit{x}}}_{2}) }$ of a pair of joint states
${\textbf{\textit{x}}}_{1}$ and ${\textbf{\textit{x}}}_{2}$ is available,
it can be used to define an MCMC algorithm (see Section \ref{XRef-Subsubsection-4611524})
for hopping around between the various joint states $\textbf{\textit{x}}$,
and which is designed to visit each joint state with a frequency
that is propartional to Pr$(\textbf{\textit{x}})$, as is required
for computing a numerical estimate of $\langle \textbf{\textit{S}}\rangle
$ in Equation \ref{XRef-Equation-322151345}.
\subsection{Markov Chain Monte Carlo Algorithm}\label{XRef-Subsubsection-4611524}

It is possible to construct an MCMC algorithm for hopping between
joint states of an MRF that respects their relative probability
of occurrence. It is {\itshape not} trivially obvious how to design
a hopping algorithm with these properties, because one has to consider
the net effect of {\itshape all} of the ways that one's proposed
algorithm can hop in to and out of each state, and to check that
this does indeed give rise to the correct joint $\Pr ( \textbf{\textit{x}})
$.

Consider a network of nodes whose joint state of its nodes splits
into two parts $(\textbf{\textit{x}},\textbf{\textit{y}})$ whose
joint probability is $\Pr ( \textbf{\textit{x}},\textbf{\textit{y}})
$. This joint probability can be split into two parts as
\begin{equation}
\Pr ( \textbf{\textit{x}},\textbf{\textit{y}}) =\Pr ( \textbf{\textit{x}}|\textbf{\textit{y}})
\Pr ( \textbf{\textit{y}}) %
\label{XRef-Equation-47114519}
\end{equation}

\noindent where $\Pr ( \textbf{\textit{x}}|\textbf{\textit{y}})
$ and $\Pr ( \textbf{\textit{y}}) $ are obtained from $\Pr ( \textbf{\textit{x}},\textbf{\textit{y}})
$ as $\Pr ( \textbf{\textit{x}}|\textbf{\textit{y}}) \equiv \frac{\Pr
( \textbf{\textit{x}},\textbf{\textit{y}}) }{\sum _{\textbf{\textit{x}}}\Pr
( \textbf{\textit{x}},\textbf{\textit{y}}) }$ and $\Pr ( \textbf{\textit{y}})
\equiv \sum _{\textbf{\textit{x}}}\Pr ( \textbf{\textit{x}},\textbf{\textit{y}})
$. Now update the joint state using $(\textbf{\textit{x}},\textbf{\textit{y}})\overset{\Pr
( {\textbf{\textit{x}}}^{\prime }|\textbf{\textit{y}}) }{\longrightarrow
}({\textbf{\textit{x}}}^{\prime },\textbf{\textit{y}})$ where ${\textbf{\textit{x}}}^{\prime
}$ is a sample that is drawn from $\Pr ( {\textbf{\textit{x}}}^{\prime
}|\textbf{\textit{y}}) $, where $\Pr ( {\textbf{\textit{x}}}^{\prime
}|\textbf{\textit{y}}) $ is a conditional probability that has the
{\itshape same} dependence on its arguments as $\Pr ( \textbf{\textit{x}}|\textbf{\textit{y}})
$ above. The joint probability $\Pr ( {\textbf{\textit{x}}}^{\prime
},\textbf{\textit{y}}) $ of the updated joint state is then
\begin{equation}
\Pr ( {\textbf{\textit{x}}}^{\prime },\textbf{\textit{y}}) =\Pr
( {\textbf{\textit{x}}}^{\prime }|\textbf{\textit{y}}) \Pr ( \textbf{\textit{y}})
\label{XRef-Equation-47114526}
\end{equation}

\noindent Comparing Equation \ref{XRef-Equation-47114519} with Equation
\ref{XRef-Equation-47114526} shows that the new joint probability
$\Pr ( {\textbf{\textit{x}}}^{\prime },\textbf{\textit{y}}) $ is
the {\itshape same} function of its arguments as the old joint probability
$\Pr ( \textbf{\textit{x}},\textbf{\textit{y}}) $, by construction.
This would {\itshape not }be the case if the sample ${\textbf{\textit{x}}}^{\prime
}$ was drawn from a $\Pr ( {\textbf{\textit{x}}}^{\prime }|\textbf{\textit{y}})
$ that did {\itshape not} have the same dependence on its arguments
as $\Pr ( \textbf{\textit{x}}|\textbf{\textit{y}}) $ above.

The above argument shows that if you have a network whose joint
probability is $\Pr ( \textbf{\textit{x}},\textbf{\textit{y}}) $,
and assuming that the network starts in an initial joint state $(\textbf{\textit{x}},\textbf{\textit{y}})$
that has joint probability $\Pr ( \textbf{\textit{x}},\textbf{\textit{y}})
$, then updating the joint state using $(\textbf{\textit{x}},\textbf{\textit{y}})\overset{\Pr
( {\textbf{\textit{x}}}^{\prime }|\textbf{\textit{y}}) }{\longrightarrow
}({\textbf{\textit{x}}}^{\prime },\textbf{\textit{y}})$ guarantees
that the new joint state $({\textbf{\textit{x}}}^{\prime },\textbf{\textit{y}})$
has joint probability $\Pr ( {\textbf{\textit{x}}}^{\prime },\textbf{\textit{y}})
$ (which has the same dependence on its arguments as $\Pr ( \textbf{\textit{x}},\textbf{\textit{y}})
$). Thus the joint probability of the joint state of the network
nodes maps to itself under the update prescription $(\textbf{\textit{x}},\textbf{\textit{y}})\overset{\Pr
( {\textbf{\textit{x}}}^{\prime }|\textbf{\textit{y}}) }{\longrightarrow
}({\textbf{\textit{x}}}^{\prime },\textbf{\textit{y}})$.

Typically, a \textit{sequence} of updates is applied, where the
joint state of the network is split into two parts in {\itshape
different} ways for successive updates, so that eventually all the
nodes in the network are visited for updating. The overall effect
is that updating causes the network to move around in the joint
state space of its nodes, whilst guaranteeing that the joint probability
of the network node states stays the same.

On the other hand, if the initial joint state $(\textbf{\textit{x}},\textbf{\textit{y}})$
does \textit{not} have joint probability $\Pr ( \textbf{\textit{x}},\textbf{\textit{y}})
$, then $\Pr ( {\textbf{\textit{x}}}^{\prime },\textbf{\textit{y}})
$ and $\Pr ( \textbf{\textit{x}},\textbf{\textit{y}}) $ will {\itshape
not} be the same functions of their arguments, so the joint probability
will \textit{change} as the updating scheme is applied. If a sequence
of updates (using a variety of splittings of the network of nodes,
as described above) is applied then this evolution can converge
to a fixed point where the joint probability is \textit{stationary}
under updating. However, convergence to a unique fixed point is
not actually guaranteed, because an inappropriate update prescription
could be used that leads to non-ergodic behaviour where the whole
joint state space is not explored, for instance. However, in practical
problems with soft joint probabilities convergence usually occurs.

In an MRF the ratio of conditional probabilities $\frac{\Pr ( {{\textbf{\textit{x}}}^{\prime
}}_{1}|\textbf{\textit{y}}) }{\Pr ( {{\textbf{\textit{x}}}^{\prime
}}_{2}|\textbf{\textit{y}}) }$ that is used to generate the MCMC
updates $(\textbf{\textit{x}},\textbf{\textit{y}})\overset{\Pr (
{\textbf{\textit{x}}}^{\prime }|\textbf{\textit{y}}) }{\longrightarrow
}({\textbf{\textit{x}}}^{\prime },\textbf{\textit{y}})$ is given
in Equation \ref{XRef-Equation-47123935}. If the joint states ${{\textbf{\textit{x}}}^{\prime
}}_{1}$ and ${{\textbf{\textit{x}}}^{\prime }}_{2}$ differ in only
a few of their vector components, then there is a lot of cancellation
in $\frac{\Pr ( {{\textbf{\textit{x}}}^{\prime }}_{1}|\textbf{\textit{y}})
}{\Pr ( {{\textbf{\textit{x}}}^{\prime }}_{2}|\textbf{\textit{y}})
}$ so the fully simplified expression for $\frac{\Pr ( {{\textbf{\textit{x}}}^{\prime
}}_{1}|\textbf{\textit{y}}) }{\Pr ( {{\textbf{\textit{x}}}^{\prime
}}_{2}|\textbf{\textit{y}}) }$ is relatively simple. This is what
makes MCMC algorithms so appropriate for MRF networks.
\subsection{Inference Using an MRF}\label{XRef-Subsubsection-4712833}

Image processing is an area where MRFs have proved to be particularly
useful \cite{GemanGeman1984}. The starting point is to define an
MRF model of the joint probability $\Pr ( \textbf{\textit{x}}) $
of the image pixels
\begin{equation}
\begin{array}{rl}
 \Pr ( \textbf{\textit{x}})  & \equiv \sum _{\textbf{\textit{y}}}\Pr
( \textbf{\textit{x}},\textbf{\textit{y}})  \\
 \Pr ( \textbf{\textit{x}},\textbf{\textit{y}})  & =\Pr ( \textbf{\textit{x}}|\textbf{\textit{y}})
\Pr ( \textbf{\textit{y}})
\end{array}
\end{equation}

\noindent where $\Pr ( \textbf{\textit{x}}) $ is expressed as the
marginal probability of $\Pr ( \textbf{\textit{x}},\textbf{\textit{y}})
$ after the hidden variables $\textbf{\textit{y}}$ have been averaged
over, and both $\Pr ( \textbf{\textit{x}}|\textbf{\textit{y}}) $
and $\Pr ( \textbf{\textit{y}}) $ may be written as products of
factors using the HCE in Equation \ref{XRef-Equation-322152418}.
The hidden variables $\textbf{\textit{y}}$ are the unobserved causes
that determine the values of the image pixels $\textbf{\textit{x}}$,
and are thus the causal factors that are used to construct a generative
model of the image. This generative model can be multi-layered with
several levels of hidden variables.

To compute the probability of the joint state of the hidden variables
$\textbf{\textit{y}}$ given an observation of the image pixel values
$\textbf{\textit{x}}$ the posterior probability $\Pr ( \textbf{\textit{y}}|\textbf{\textit{x}})
$ must be used, which may be obtained using Bayes theorem as
\begin{equation}
\Pr ( \textbf{\textit{y}}|\textbf{\textit{x}}) =\frac{\Pr ( \textbf{\textit{x}}|\textbf{\textit{y}})
\Pr ( \textbf{\textit{y}}) }{\sum _{\textbf{\textit{y}}}\Pr ( \textbf{\textit{x}}|\textbf{\textit{y}})
\Pr ( \textbf{\textit{y}}) }%
\label{XRef-Equation-322153510}
\end{equation}

\noindent An MCMC algorithm (see Section \ref{XRef-Subsubsection-4611524})
can then be used to draw samples from $\Pr ( \textbf{\textit{y}}|\textbf{\textit{x}})
$. Note that successive samples produced by the MCMC algorithm are
strongly correlated with each other because the MCMC algorithm has
a finite memory time; this makes MCMC run times (for a given size
of error bar) much longer than would be the case if the samples
could be somehow independently drawn from $\Pr ( \textbf{\textit{y}}|\textbf{\textit{x}})
$.

Also, if $\Pr ( \textbf{\textit{y}}|\textbf{\textit{x}}) $ has a
{\itshape single} well-defined peak of probability, then the MCMC
algorithm can be used to locate this, usually with the assistance
of a simulated annealing algorithm to ``soften'' $\Pr ( \textbf{\textit{y}}|\textbf{\textit{x}})
$ during the early stages of the algorithm, and then MCMC fluctuations
about this peak can be observed in order to deduce the robustness
of the solution.

Typically, in image processing applications there is a single overwhelmingly
likely hidden variables interpretation of the image pixels (i.e.
$\Pr ( \textbf{\textit{y}}|\textbf{\textit{x}}) $ has a {\itshape
single} well-defined peak of probability). However, the above approach
gracefully (and consistently) degrades when the interpretation is
ambiguous (i.e. $\Pr ( \textbf{\textit{y}}|\textbf{\textit{x}})
$ does {\itshape not} have a single well-defined peak of probability).
This graceful degradation in the face of ambiguity is one of the
strengths of the Bayesian approach.
\subsection{Multiply Occupied States}\label{XRef-Subsubsection-331134331}

It is useful to develop a concrete way of visualising the hopping
processes that underlie the MCMC algorithm described in Section
\ref{XRef-Subsubsection-4611524}. This is a prerequisite for the
generalisation of MCMC algorithms that developed later in Section
\ref{XRef-Subsection-33113632}.

The state $\textbf{\textit{x}}$ of an $N$-node MRF is $\textbf{\textit{x}}\equiv
(x_{1},x_{2},\cdots ,x_{N})$, and for a given $\textbf{\textit{x}}$
each of its components $x_{i}$ lives in {\itshape one} of an assumed
finite number $m$ of states that are available to $x_{i}$, where
for simplicity we assume that all the $x_{i}$ have the \textit{same}
number of states $m$. One way of representing each $x_{i}$ is as
an $m$-component vector $(0,0,\cdots ,0,1,0,\cdots ,0,0)$, where
the ``1'' identifies which of the $m$ states $x_{i}$ happens to
have. This representation is essentially a histogram with $m$ bins,
with a {\itshape single} sample occupying one of the bins. The whole
state of the $N$ component $\textbf{\textit{x}}$ vector is then
represented by $N$ such histograms, each with a {\itshape single}
``1'' placed in the appropriate bin to identify the state of \textit{all}
of the $x_{i}$ for $i=1,2,\cdots ,N$. Naturally, this use of histograms
is an exceedingly wasteful coding of the state $\textbf{\textit{x}}$
because it consists mostly of ``0'' entries. However, it {\itshape
does} allow the hopping operations that are generated by the MCMC
algorithm to be represented directly as operations in which each
``1'' hops around between the bins of its histogram. More importantly,
this representation of the MRF state is suitable for the generalisation
in Section \ref{XRef-Subsection-33113632} where each histogram will
have \textit{multiple} samples occupying its bins (i.e. multiple
states will be recorded at each MRF node). This is discussed in
more detail below.
\begin{figure}[h]
\begin{center}
\includegraphics{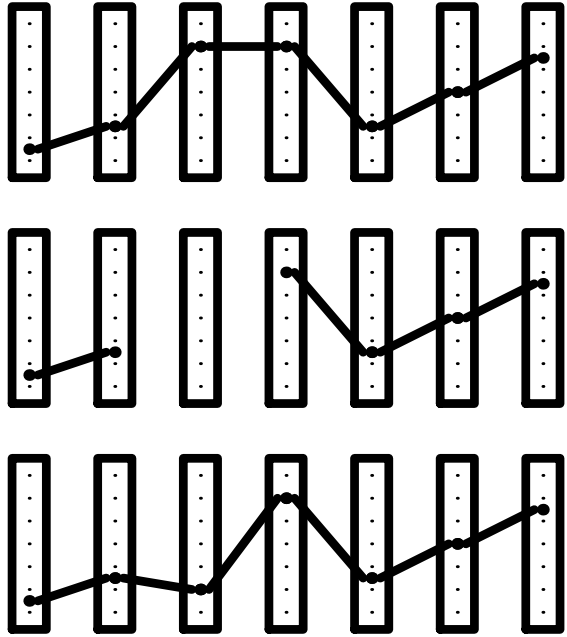}

\end{center}
\caption{Steps of an MCMC update of a Markov chain with $N=7$ and
$m=7$.}\label{XRef-FigureCaption-101810312}
\end{figure}

Figure \ref{XRef-FigureCaption-101810312} shows a Markov chain with
7 nodes (i.e. $N=7$), each of which has 7 possible states (i.e.
$m=7$). The state space of each node is represented by one of the
rectangles, the particular bin that is occupied by a sample is shown
as a blob (the unoccupied bins are shown as dots), and the particular
2-clique interactions (see Equation \ref{XRef-Equation-322152418})
that are activated by the occupied node states are shown as bold
lines.
\begin{enumerate}
\item The top row of Figure \ref{XRef-FigureCaption-101810312} shows
a random initial state of the Markov chain.
\item The middle row of Figure \ref{XRef-FigureCaption-101810312}
shows that the sample in node 3 has been annihilated. This is the
\textit{first} step of an MCMC update, in which a node is chosen
at random and its state is erased.\label{XRef-Item1Numbered-1018101151}
\item The bottom row of Figure \ref{XRef-FigureCaption-101810312}
shows that a sample in node 3 has been created. This is the \textit{second}
step of an MCMC update, in which a sample is created in node 3 whose
state was previously erased in step \ref{XRef-Item1Numbered-1018101151}
above. The influence of the neighbouring nodes is used to probabilistically
determine the state in which to create the sample, as described
in Section \ref{XRef-Subsubsection-4611524}.
\begin{figure}[h]
\begin{center}
\includegraphics{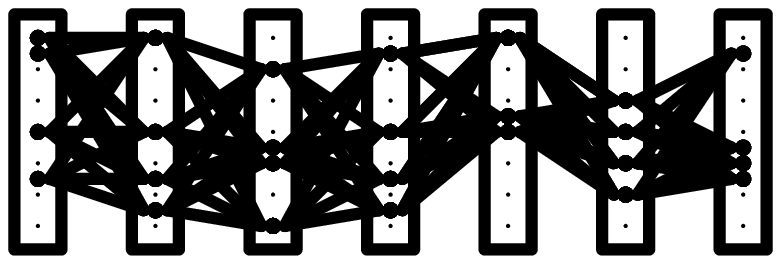}

\end{center}
\caption{Multiply occupied Markov chain showing a random state.}\label{XRef-FigureCaption-1018103048}
\end{figure}
\end{enumerate}

The histogram representation allows generalisations of the MCMC
algorithm in which each MRF node is occupied by more than one sample,
when it is said to be multiply occupied. Figure \ref{XRef-FigureCaption-1018103048}
shows an example of this type of MRF state.

It is important not to confuse multiply occupied states with other
uses of state space:
\begin{enumerate}
\item Histograms with more than one sample are {\itshape not} the
same as ensembles of histograms each with one sample. This is because
the former allow for the possibility that the MCMC algorithm can
cause the samples to interact with each other, whereas the latter
is a means of running multiple standard MCMC algorithms in parallel.
\item Histograms with more than one sample could be viewed as having
a single ``super''-state that recorded as a single state the entire
contents of the histogram bins, which would disguise the fact that
the histogram was actually constructed out of samples occupying
the histogram bins. The higher level super-state description is
mathematically equivalent to the lower-level description in terms
of individual samples, but it does \textit{not} allow the development
of\ \ detailed MCMC algorithms. We prefer to view the higher level
super-state description as an interpretation that is used \textit{after}
the lower level details have been worked out using the techniques
that are presented in this paper.
\end{enumerate}

In Figure \ref{XRef-FigureCaption-1018103048} the histogram associated
with each node contains more than one sample. Such multiple occupancy
was {\itshape not} present in the basic MRF theory of Section \ref{XRef-Subsubsection-331132456},
so the detailed form of the MCMC algorithm of Section \ref{XRef-Subsubsection-4611524}
must now be generalised. Multiple occupancy is explored in detail
in Section \ref{XRef-Subsection-33113632} using creation and annihilation
operator techniques to hop samples between histogram bins, which
is achieved by annihilating a sample from one bin and creatng a
sample in another bin, as illustrated in Figure \ref{XRef-FigureCaption-101810312}.

When more than one sample per histogram is allowed then various
new types of processing become possible:
\begin{enumerate}
\item The number of samples per histogram can be varied with time.
This requires birth and death rules as well as migration (or hopping)
rules for the histogram samples. In this case the creation and annihilation
operators would be applied in ways that do \textit{not} enforce
conservation of the number of samples in each histogram, so annihilation
without subsequent creation (and vice versa) are permitted operations.
This is how ``reversible jump'' MCMC algorithms \cite{Green1995}
might be implemented using creation and annihilation operators.
\item The samples can interact with each other in complicated ways
to form ``bound states'', which would then behave like higher level
``symbols'' (i.e. sets of interacting histogram samples) that are
constructed out of ``sub-symbols'' (i.e. the histogram samples themselves).
This is illustrated in Figure \ref{XRef-FigureCaption-46141421},
Figure \ref{XRef-FigureCaption-46142153} and Figure \ref{XRef-FigureCaption-11520345}
below.
\begin{figure}[h]
\begin{center}
\includegraphics{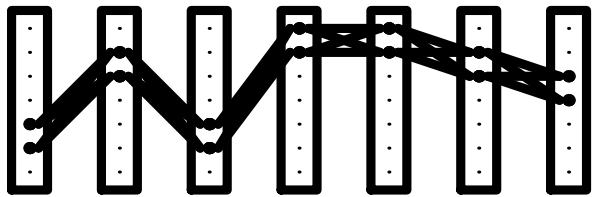}

\end{center}
\caption{Multiply occupied Markov chain showing a tube-like joint
state.}\label{XRef-FigureCaption-46141421}
\end{figure}
\end{enumerate}

Figure \ref{XRef-FigureCaption-46141421} shows a multiple-sample
version of Figure \ref{XRef-FigureCaption-101810312} that is more
highly structured than the example shown in Figure \ref{XRef-FigureCaption-1018103048}.
For illustrative purposes, the samples are now assumed to be in
neighbouring states at each node rather than spread out at random;
typically this would be the case for Markov chains whose properties
are optimised to encode information in a topographically ordered
way. The 2-cliques that then contribute typically form the tube-like
joint state of activated 2-cliques shown in Figure \ref{XRef-FigureCaption-46141421}.
\begin{figure}[h]
\begin{center}
\includegraphics{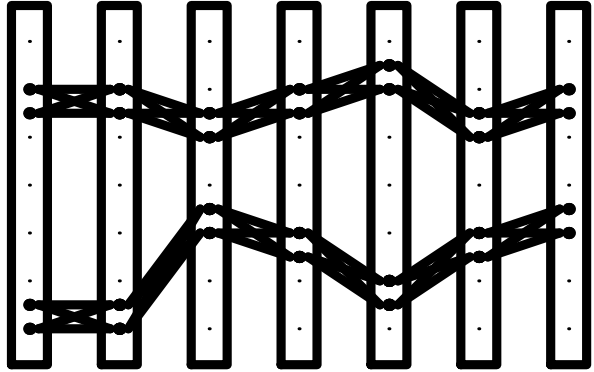}

\end{center}
\caption{Multiply occupied Markov chain showing two parallel tube-like
joint states.}\label{XRef-FigureCaption-46142153}
\end{figure}

Figure \ref{XRef-FigureCaption-46142153} shows another possibility
that can arise with multiple sample occupancy, where the occupancy
of each node splits into two separate clusters of samples, {\itshape
and} where the probability factors associated with the 2-cliques
is such that only node states that are both in the top half of the
diagram are connected (and similarly for the bottom half of the
diagram), so that there are no activated 2-cliques running between
the top and bottom halves of the diagram (or at least the contribution
of these is negligible). Effectively, this multiply occupied Markov
chain has two completely independent Markov chains embedded within
it, each of which has its own tube-like joint state of activated
2-cliques. This type of structure emerges in multiply occupied Markov
chains that have a \textit{limited} number of states available to
each node of the chain, and which are optimised to encode information
topographically (which ensures that the tube-like joint states are
localised in the node state spaces). This\ \ type of behaviour emerges
when\ \ SON training methods are used, but it will not be discussed
further in this paper.
\begin{figure}[h]
\begin{center}
\includegraphics{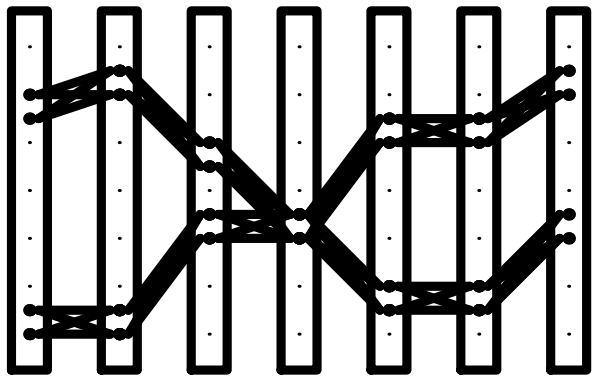}

\end{center}
\caption{Multiply occupied Markov chain showing two parallel ``tube''
states bound together.}\label{XRef-FigureCaption-11520345}
\end{figure}

Figure \ref{XRef-FigureCaption-11520345} shows how Figure \ref{XRef-FigureCaption-46142153}
can be modified if the two tube-like joint states have some node
states in common, which\ \ binds the tubes together. An extreme
version of this binding between tubes can occur if the situation
is as shown in Figure \ref{XRef-FigureCaption-46142153}, but \textit{additionally}
there are some weak interactions between the tubes.
\section{Operator Implementation of MCMC Algorithms}\label{XRef-Subsection-33113632}

The aim of this section is to present a theoretical framework for
expressing MCMC algorithms, which is based on\ \ operators that
have very simple algebraic properties, but which is nevertheless
sufficiently flexible that it allows a large class of MCMC-like
algorithms to be represented.

Section \ref{XRef-Subsection-1021112317} gives some background material
that motivates the use of MCMC algorithms as the primary means of
building dynamical models for discrete networks. Section \ref{XRef-Subsubsection-331145022}
introduces creation and annihilation operators for manipulating
samples in multiply occupied network nodes. Section \ref{XRef-Subsection-322174917}
uses these basic operators to construct a composite operator for
generating MCMC updates, Finally, Section \ref{XRef-Subsubsection-91313212}
summarises a diagrammatic representation of MCMC algorithms.
\subsection{Background}\label{XRef-Subsection-1021112317}

The aim here is to rewrite the MCMC algorithm for running an MRF
(see Section \ref{XRef-Subsubsection-4611524}) using operator algebra.
This will allow the algorithm to be run in state spaces where the
basic MCMC algorithm has not previously been used, and will thus
generalise the algorithm. Throughout this section the emphasis is
on using the MCMC algorithm as the \textit{starting point} for deducing
the properties of an MRF, so the MRF is viewed as corresponding
to the equilibrium behaviour of a (stochastic) discrete-time dynamical
system. Hitherto, the MCMC algorithm could be viewed as an artefact
of a particular way of sampling from an MRF, but here it is viewed
as the way in which the MRF actually behaves. This moves slightly
away from the original motivation for using MRFs to model and manipulate
joint probabilities for use in Bayesian calculations (see Section
\ref{XRef-Section-116135347}), but this change of emphasis allows
full advantage to taken of the flexibility of the MCMC approach,
and in particular its generalisation to multiply occupied states.

This jump to using discrete-time dynamical systems as the starting
point for building models allows a much larger class of behaviours
to be explored, including ones that do \textit{not} have a corresponding
HCE representation of the equilibrium behaviour (i.e. as a simple
product of probability factors, as in Equation \ref{XRef-Equation-322152418}),
or do \textit{not} have a steady state equilibrium behaviour at
all (e.g. a limit \textit{cycle} rather than a limit \textit{point},
etc).

The MCMC approach models everything as part of a dynamical evolution
process, where a static statistical model of the world is obtained
by taking a snapshot of the evolution of the dynamical system. Those
who insist on starting from a fixed graphical model based on the
HCE (or a set of such models) might be disappointed that this is
\textit{not} the starting point that is used here. However, they
should note that the underlying process that generates their graphical
model in the first place is actually dynamical, and that their model
merely describes the statistical properties through a time slice
of this dynamical process; in other words, their model describes
only a \textit{marginal} distribution. For instance, an MRF image
model does \textit{not} attempt to model the history of the dynamical
processes that cause the (hidden) objects to eventually give rise
to the observed pixel values. Analogously, \textit{all} MRFs derive
from a hidden dynamical process.

The results presented in this section make use of creation and annihilation
operator techniques to generate the hopping processes that underlie
MCMC algorithms, which allows MCMC algorithms to be written using
a very compact notation. These operator techniques will be familiar
to physicists who use quantum field theory (QFT) \cite{Zee2003},
and for the convenience of physicists the notation used here is
the same as is used in QFT. Generally, creation and annihilation
operators can be used to generate birth and death processes (respectively),
which thus increase and decrease the dimensionality of the state
space (respectively), so this approach naturally lends itself to
describing processes that correspond to ``reversible jump'' MCMC
algorithms \cite{Green1995}.
\subsection{Creation and Annihilation Operators}\label{XRef-Subsubsection-331145022}

In this section the mathematical development of the properties of
creation and annihilation operators is deliberately presented in
an informal way, by expressing it in terms of operations on the
samples occupying histogram bins. This is to encourage a concrete
and intuitive understanding of how these operators act on samples,
rather than to merely think of them as objects that have particular
algebraic properties. To a physicist who is familiar with the use
of these techniques in QFT, the explanations will appear to be very
long-winded and the derivations very cavalier, and to them we apologise.
\subsubsection{Multiply Occupied States}

The multiply occupied states described in Section \ref{XRef-Subsubsection-331134331}
can be manipulated by suitably defined creation and annihilation
operators.

Multiply occupied states can viewed as hsistograms with multiple
samples occupying the histogram bins. These histograms can be represented
thus:
\begin{enumerate}
\item Empty histogram: $|0\rangle $. This represents the bins (an
indeterminate number of them) of a histogram with no samples in
any of the bins. The notation $|0\rangle $ has been chosen to correspond
exactly to the ``vacuum'' state as used by physicists; it represents
the background in which we will create and annihilate histogram
samples (or particles).\label{XRef-Item1Numbered-331134841}
\item Histogram with one sample in bin $i$: ${a_{i}}^{\dagger }|0\rangle
$. The $|0\rangle $ represents the empty histogram (as defined above),
and the creation operator ${a_{i}}^{\dagger }$ acting from the left
represents the action of creating one sample in bin $i$ of the empty
histogram. The notation ${a_{i}}^{\dagger }$ has been chosen to
correspond exactly to the operator for creating a particle in state
$i$ as used by physicists, and the notation ${a_{i}}^{\dagger }|0\rangle
$ corresponds exactly to the notation for a single particle in state
$i$. The use of the dagger notation $\dagger $ (i.e. adjoint operator)
is chosen to make our notation compatible with that used in QFT
\cite{Zee2003}, which will be discussed in more detail in Section
\ref{XRef-Subsubsection-116143013}.\label{XRef-Item1Numbered-331135457}
\item Histogram with $n_{i}$ samples in bin $i$ : ${({a_{i}}^{\dagger
})}^{n_{i}}|0\rangle $. This is a multiply occupied histogram, which
is obtained by operating on the empty histogram $|0\rangle $ multiple
times with the creation operator ${a_{i}}^{\dagger }$.
\item Histogram with $n_{i}$ samples in bin $i$ (for $i=1,2,\cdots
,m$): $\prod _{i=1}^{m}{({a_{i}}^{\dagger })}^{n_{i}}|0\rangle $.
This is a straightforward generalisation of the above, where creation
operators are applied multiple times to all of the histogram bins.
\end{enumerate}

The above representation of histogram states does \textit{not} provide
a means for freely manipulating them. In order to be able to do
this it is necessary to be able to annihilate samples as well as
create them as above.
\subsubsection{Creation and Annihilation Operators}\label{XRef-Subsubsection-11713637}

The annihilation operations discussed below may be achieved by using
the annihilation operator $a_{i}$ which is the adjoint of the creation
operator ${a_{i}}^{\dagger }$. See the discussion on adjoint operators
in Section \ref{XRef-Subsubsection-116143013} for more details on
why the creation operator ${a_{i}}^{\dagger }$ and annihilation
operator $a_{i}$ are adjoints of each other. Note that in the description
immediately below the behaviour of ${a_{i}}^{\dagger }$ and $a_{i}$
corresponds to our intuitive notion of how these operators should
behave, rather than formally derived from their algebraic properties
which are presented later on in Section \ref{XRef-Subsubsection-1018145547}.

Annihilating a sample from an empty histogram erases the state space
itself. This simply {\itshape defines} what happens when you try
to remove a sample from an already empty histogram, which is very
useful for cleaning up algebraic expressions involving $a_{i}$ and
$|0\rangle $. In effect, this defines the ``vacuum'' $|0\rangle
$ as the reference state for determining the occupancy of each histogram
bin.
\begin{equation}
a_{i}\left| 0\right\rangle  =0%
\label{XRef-Equation-32216231}
\end{equation}

\noindent which can be represented for a 4-bin histogram for any
$i$ as
\begin{equation}
\begin{array}{ccc}
 \left( 0,0,0,0\right)  & \overset{a_{i}}{\longrightarrow } & 0
\end{array}
\end{equation}

Annihilating a sample from a 1-sample histogram leaves an empty
histogram. This definition is the common-sense notion of what should
happen when you create a sample in a histogram bin, then annihilate
it again. Thus
\begin{equation}
a_{i}{a_{i}}^{\dagger }\left| 0\right\rangle  =\left| 0\right\rangle
\label{XRef-Equation-46163415}
\end{equation}

\noindent which can be represented for a 4-bin histogram and for
$i=3$ as
\begin{equation}
\begin{array}{ccccc}
 \left( 0,0,0,0\right)  & \overset{{a_{i}}^{\dagger }}{\longrightarrow
} & \left( 0,0,1,0\right)  & \overset{a_{i}}{\longrightarrow } &
\left( 0,0,0,0\right)
\end{array}%
\label{XRef-Equation-4714300}
\end{equation}

Annihilating the {\itshape wrong} sample (i.e. $j\neq i$) from a
1-sample histogram erases the state space itself. This is a generalisation
of Equation \ref{XRef-Equation-32216231} in which the histogram
already contains one sample, but it is in a \textit{different} bin
from the one from which we are trying to remove a sample.
\begin{equation}
\begin{array}{ccc}
 \left. a_{j}{a_{i}}^{\dagger }\left| 0\right. \right\rangle  =0
&   & j\neq i
\end{array}%
\label{XRef-Equation-46163424}
\end{equation}

\noindent which can be represented for a 4-bin histogram and for
$i=3$ and $j\neq i$ as
\begin{equation}
\begin{array}{ccccc}
 \left( 0,0,0,0\right)  & \overset{{a_{i}}^{\dagger }}{\longrightarrow
} & \left( 0,0,1,0\right)  & \overset{a_{j}}{\longrightarrow } &
0
\end{array}%
\label{XRef-Equation-47143011}
\end{equation}

\noindent Equation \ref{XRef-Equation-4714300} and Equation \ref{XRef-Equation-47143011}
can now be combined to give (the illustration shows the $i=3$ case)
\begin{equation}
\begin{array}{ccccc}
 \left( 0,0,0,0\right)  & \overset{{a_{i}}^{\dagger }}{\longrightarrow
} & \left( 0,0,1,0\right)  & \overset{a_{j}}{\longrightarrow } &
\begin{array}{cc}
 \left( 0,0,0,0\right)  & j=i \\
 0 & j\neq i
\end{array}
\end{array}
\end{equation}

If the location of the occupied bin is unknown, yet you want to
be certain that you annihilate the sample, then you have to attempt
to annihilate a sample from every one of the histogram bins. This
combines the properties of both Equation \ref{XRef-Equation-46163415}
and Equation \ref{XRef-Equation-46163424}. Note that $|0\rangle
$ (the empty histogram) is different from $0$ (no histogram at all,
i.e. not even an empty one).
\begin{equation}
\begin{array}{rl}
 \left. \left( \sum _{j=1}^{m}a_{j}\right)  {a_{i}}^{\dagger } \left|
0\right. \right\rangle   & =\left. \left. \left. \left. a_{1}{a_{i}}^{\dagger
} \left| 0\right. \right\rangle  +a_{2}{a_{i}}^{\dagger } \left|
0\right. \right\rangle  +\cdots +a_{i}{a_{i}}^{\dagger }\left| 0\right.
\right\rangle  +\cdots +a_{m}{a_{i}}^{\dagger } \left| 0\right.
\right\rangle   \\
  & \left. =0+0+\cdots +0+\left| 0\right. \right\rangle  +0+\cdots
+0 \\
  & \left. =\left| 0\right. \right\rangle
\end{array}%
\label{XRef-Equation-46163523}
\end{equation}

\noindent which can be represented for a 4-bin histogram and for
$i=3$ as
\begin{equation}
\begin{array}{ccccc}
 \left( 0,0,0,0\right)  & \overset{{a_{i}}^{\dagger }}{\longrightarrow
} & \left( 0,0,1,0\right)  & \overset{\sum _{j=1}^{m}a_{j}}{\longrightarrow
} & \left( 0,0,0,0\right)
\end{array}
\end{equation}

Annihilating a sample from a 2-sample histogram (samples in {\itshape
different} bins, i.e. $i_{1}\neq i_{2}$) leaves two 1-sample histograms.
This is a generalisation of Equation \ref{XRef-Equation-46163523}
in which the histogram starts with {\itshape two} samples (known
to be in different bins) rather than {\itshape one} sample.
\begin{equation}
\begin{array}{rl}
 \left. \left( \sum _{j=1}^{m}a_{j}\right)  {a_{i_{1}}}^{\dagger
}{a_{i_{2}}}^{\dagger } \left| 0\right. \right\rangle   & =\begin{array}{c}
 \left. \left. a_{1}{a_{i_{1}}}^{\dagger }{a_{i_{2}}}^{\dagger }
\left| 0\right. \right\rangle  +\cdots +a_{i_{i}}{a_{i_{1}}}^{\dagger
}{a_{i_{2}}}^{\dagger }\left| 0\right. \right\rangle  +\cdots  \\
 \left. \left. \cdots +a_{i_{2}}{a_{i_{1}}}^{\dagger }{a_{i_{2}}}^{\dagger
} \left| 0\right. \right\rangle  +\cdots +a_{m}{a_{i_{1}}}^{\dagger
}{a_{i_{2}}}^{\dagger } \left| 0\right. \right\rangle
\end{array} \\
  & =\begin{array}{c}
 \left. 0+\cdots +0+{a_{i_{2}}}^{\dagger } \left| 0\right. \right\rangle
+0+\cdots  \\
 \left. \cdots +0+{a_{i_{1}}}^{\dagger } \left| 0\right. \right\rangle
+0+\cdots +0
\end{array} \\
  & =\begin{array}{ccc}
 \left. \left. {a_{i_{1}}}^{\dagger } \left| 0\right. \right\rangle
+{a_{i_{2}}}^{\dagger } \left| 0\right. \right\rangle   &   & i_{1}\neq
i_{2}
\end{array}
\end{array}%
\label{XRef-Equation-46164216}
\end{equation}

\noindent which can be represented for a 4-bin histogram and for
$(i_{1},i_{2})=(1,3)$ as
\begin{equation}
\begin{array}{ccccccc}
 \left( 0,0,0,0\right)  & \overset{{a_{i_{1}}}^{\dagger }}{\longrightarrow
} & \left( 1,0,0,0\right)  & \overset{{a_{i_{2}}}^{\dagger }}{\longrightarrow
} & \left( 1,0,1,0\right)  & \overset{\sum _{j=1}^{m}a_{j}}{\longrightarrow
} & \begin{array}{c}
 \left( 1,0,0,0\right)  \\
 + \\
 \left( 0,0,1,0\right)
\end{array}
\end{array}
\end{equation}

Annihilating a sample from a 2-sample histogram (samples in the
{\itshape same} bin, i.e. $i_{1}=i_{2}=i$) leaves two copies of
the same 1-sample histogram (because either of the two samples can
be annihilated to leave one sample). This is a variation of Equation
\ref{XRef-Equation-46164216}, and it is the \textit{first} example
of attempting to annihilate a sample from a bin that has more than
one sample in it. The number of ways of annihilating a sample from
a multiply occupied bin is equal to the number of samples in the
bin.
\begin{equation}
\begin{array}{rl}
 \left. \left( \sum _{j=1}^{m}a_{j}\right)  {\left( {a_{i}}^{\dagger
}\right) }^{2}\left| 0\right. \right\rangle   & =\begin{array}{c}
 \left. \left. a_{1} {\left( {a_{i}}^{\dagger }\right) }^{2} \left|
0\right. \right\rangle  +a_{2} {\left( {a_{i}}^{\dagger }\right)
}^{2} \left| 0\right. \right\rangle  +\cdots  \\
 \left. \left. \cdots +a_{i} {\left( {a_{i}}^{\dagger }\right) }^{2}\left|
0\right. \right\rangle  +\cdots +a_{m} {\left( {a_{i}}^{\dagger
}\right) }^{2} \left| 0\right. \right\rangle
\end{array} \\
  & =\left. 0+0+\cdots +0+2{a_{i}}^{\dagger }\left| 0\right. \right\rangle
+0+\cdots +0 \\
  & =\left. 2{a_{i}}^{\dagger } \left| 0\right. \right\rangle
\end{array}%
\label{XRef-Equation-33114267}
\end{equation}

\noindent which can be represented for a 4-bin histogram and for
$i_{1}=1$ as
\begin{equation}
\begin{array}{ccccccc}
 \left( 0,0,0,0\right)  & \overset{{a_{i_{1}}}^{\dagger }}{\longrightarrow
} & \left( 1,0,0,0\right)  & \overset{{a_{i_{1}}}^{\dagger }}{\longrightarrow
} & \left( 2,0,0,0\right)  & \overset{\sum _{j=1}^{m}a_{j}}{\longrightarrow
} & \begin{array}{c}
 \left( 1,0,0,0\right)  \\
 + \\
 \left( 1,0,0,0\right)
\end{array}
\end{array}%
\label{XRef-Equation-512102316}
\end{equation}
\subsubsection{Creation and Annihilation Operator Commutation Relations}\label{XRef-Subsubsection-1018145547}

Now that some of the required properties of creation and annihilation
operators have been established, we are in a position to guess what
their general algebraic properties should be, so that we can do
arbitrarily complicated operator manipulations on states of arbitrary
occupany.

All of the above behaviour of creation and annihilation operators
(apart from $a_{i}|0\rangle =0$ in Equation \ref{XRef-Equation-32216231})
can be summarised in the following commutation relations
\begin{equation}
\begin{array}{rl}
 a_{i}{a_{j}}^{\dagger }-{a_{j}}^{\dagger }a_{i} & =\delta _{i,j}
\\
 a_{i}a_{j}-a_{j}a_{i} & =0 \\
 {a_{i}}^{\dagger }{a_{j}}^{\dagger }-{a_{j}}^{\dagger }{a_{i}}^{\dagger
} & =0
\end{array}
\end{equation}

\noindent where $\delta _{i,j}$ is a Kronecker delta ($\delta _{i,j}=1$
if $i=j$, and $\delta _{i,j}=0$ if $i\neq j$). These commutation
relations are usually written in shorthand notation as
\begin{equation}
\begin{array}{rl}
 \left[ a_{i},{a_{j}}^{\dagger }\right]  & =\delta _{i,j} \\
 \left[ a_{i},a_{j}\right]  & =0 \\
 \left[ {a_{i}}^{\dagger },{a_{j}}^{\dagger }\right]  & =0
\end{array}%
\label{XRef-Equation-32216132}
\end{equation}

The $[a_{i},a_{j}]=0$ and $[{a_{i}}^{\dagger },{a_{j}}^{\dagger
}]=0$ commutation relations follow from the fact that a sequence
consisting solely\ \ of annihilation operators (or solely of creation
operators) has the same effect whatever the order in which the operators
appear in the sequence. However, this order independence property
vanishes when the sequence contains interleaved creation and annihilation
operators, as will be explained below.

The $[a_{i},{a_{j}}^{\dagger }]=\delta _{i,j}$ commutation relation
may be illustrated for a 4-bin {\itshape empty} histogram and for
$j=3$ as
\begin{equation}
\begin{array}{ccccc}
 \left( 0,0,0,0\right)  & \overset{{a_{j}}^{\dagger }}{\longrightarrow
} & \left( 0,0,1,0\right)  & \overset{a_{i}}{\longrightarrow } &
\begin{array}{cc}
 \left( 0,0,0,0\right)  & i=j \\
 0 & i\neq j
\end{array} \\
 \left( 0,0,0,0\right)  & \overset{a_{i}}{\longrightarrow } & 0
& \overset{{a_{j}}^{\dagger }}{\longrightarrow } & 0
\end{array}
\end{equation}

\noindent and for the general histogram as
\begin{equation}
\begin{array}{ccccc}
 \left( n_{1},n_{2},\cdots \right)  & \overset{{a_{j}}^{\dagger
}}{\longrightarrow } & \left( n_{1},n_{2},\cdots ,n_{j}+1,\cdots
\right)  & \overset{a_{i}}{\longrightarrow } & \begin{array}{cc}
 \left( n_{i}+1\right)  \left( n_{1},n_{2},\cdots \right)  & i=j
\\
 n_{i} \left( n_{1},\cdots ,n_{i}-1,\cdots ,n_{j}+1,\cdots \right)
& i\neq j
\end{array} \\
 \left( n_{1},n_{2},\cdots \right)  & \overset{a_{i}}{\longrightarrow
} & n_{i} \left( n_{1},\cdots ,n_{i}-1,\cdots \right)  & \overset{{a_{j}}^{\dagger
}}{\longrightarrow } & \begin{array}{cc}
 n_{i} \left( n_{1},n_{2},\cdots \right)  & i=j \\
 n_{i} \left( n_{1},\cdots ,n_{i}-1,\cdots ,n_{j}+1,\cdots \right)
& i\neq j
\end{array}
\end{array}%
\label{XRef-Equation-520142115}
\end{equation}

\noindent and by taking the difference of the $a_{i}{a_{j}}^{\dagger
}$ (i.e. the first line in Equation \ref{XRef-Equation-520142115}
above) and the ${a_{j}}^{\dagger }a_{i}$ (i.e. the second line in
Equation \ref{XRef-Equation-520142115} above) results above the
commutator relation $[a_{i},{a_{j}}^{\dagger }]=\delta _{i,j}$ is
correctly verified. The key result is the $i=j$ case in Equation
\ref{XRef-Equation-520142115} which has a factor $n_{i}+1$ in the
$a_{i}{a_{j}}^{\dagger }$ case and a factor $n_{i}$ in the ${a_{j}}^{\dagger
}a_{i}$ case, which arises because the number of ways of annihilating
a sample is equal to the number of samples in the histogram bin
which the annihilation operator acts upon, and this number is {\itshape
one greater} in the case where a creation operator got to act on
the bin \textit{before} the annihilation operator got its chance
to act on the same bin.

Note that the commutation relation in Equation \ref{XRef-Equation-32216132}
\textit{extends} the properties of the creation and annihilation
operators independently of the states that they act upon, so that
the operators now have specific effects on histograms with multiple
samples in multiple bins; these extended properties were not specified
in the development up as far as Equation \ref{XRef-Equation-512102316}.
Thus the particular choice of commutation relation in Equation \ref{XRef-Equation-32216132}
defines a specific set of combinatoric factors for how one can select
samples for creation and annihilation, which are described above
and which have intuitively reasonable properties.

The above properties of the creation and annihilation operators
have been justified by appealing to simple operations on the samples
in histogram bins, which leads automatically these operators having
the same combinatoric properties as the creation and annihilation
operators that are used in a QFT of bosons \cite{Zee2003}.
\subsubsection{Commutation Relations Generalise MCMC Algorithms}

In Section \ref{XRef-Subsubsection-1018145547} a set of commutation
relations was defined based on the required properties of the creation
and annihilation operators in a variety of simple cases that were
discussed in Section \ref{XRef-Subsubsection-11713637}. However,
these commutation relations do more than just summarise these special
cases, they extend the use of creation and annihilation operators
to \textit{all} situations, including cases where the histogram
bins are occcupied by an arbitrary number of samples. Thus these
commutation relations provide an algebraically simple route to generalisation
of MCMC algorithms. No doubt there are other generalisations of
the standard MCMC algorithm, but none of them will have the algebraic
simplicity of the properties defined in Section \ref{XRef-Subsubsection-1018145547}.

For instance, consider the multiply occupied state ${({a_{1}}^{\dagger
})}^{n_{1}}\cdots  {({a_{m}}^{\dagger })}^{n_{m}} |0\rangle $. As
in QFT \cite{Zee2003}, the creation operators can be used to construct
a Fock space of states with all possible occupancies, and this Fock
space can be explored by applying creation and annihilation operators.
This type of exploration corresponds to what is done in reversible
jump MCMC algorithms \cite{Green1995}, where the scope of MCMC updates
is extended so that they sample from various models, in additional
to the sampling within a single model that usually occurs.

It can be seen that the effect of $\sum _{j=1}^{m}a_{j}$ is to count
the number of samples in each histogram bin (i.e. the number of
ways of annihilating a sample from a bin is equal to the number
of samples in the bin), and to also annihilate one of the samples
from each bin, as is shown in Equation \ref{XRef-Equation-41124426}.
\begin{equation}
\left. \left( \sum _{j=1}^{m}a_{j}\right)  {\left( {a_{1}}^{\dagger
}\right) }^{n_{1}}{\left( {a_{2}}^{\dagger }\right) }^{n_{2}}\cdots
{\left( {a_{m}}^{\dagger }\right) }^{n_{m}} \left| 0\right. \right\rangle
=\begin{array}{c}
 \left. n_{1} {\left( {a_{1}}^{\dagger }\right) }^{n_{1}-1}{\left(
{a_{2}}^{\dagger }\right) }^{n_{2}}\cdots  {\left( {a_{m}}^{\dagger
}\right) }^{n_{m}} \left| 0\right. \right\rangle   \\
 \left. +n_{2} {\left( {a_{1}}^{\dagger }\right) }^{n_{1}}{\left(
{a_{2}}^{\dagger }\right) }^{n_{2}-1}\cdots  {\left( {a_{m}}^{\dagger
}\right) }^{n_{m}} \left| 0\right. \right\rangle   \\
 \vdots  \\
 \left. +n_{m} {\left( {a_{1}}^{\dagger }\right) }^{n_{1}}{\left(
{a_{2}}^{\dagger }\right) }^{n_{2}}\cdots  {\left( {a_{m}}^{\dagger
}\right) }^{n_{m}-1} \left| 0\right. \right\rangle
\end{array}%
\label{XRef-Equation-41124426}
\end{equation}

\noindent The above deficit of one sample after the application
of $\sum _{j=1}^{m}a_{j}$ can be rectified by altering the operator
as\ \ $\sum _{j=1}^{m}a_{j}\longrightarrow \sum _{j=1}^{m}{a_{j}}^{\dagger
}a_{j}$, because the inclusion of ${a_{j}}^{\dagger }$ to the left
of $a_{j}$ ensures that a sample will be created in bin $j$ to make
up for the one that $a_{j}$ annihilated. Note that there is only
{\itshape one} way of creating a sample in a bin, but there are
as many ways of annihilating a sample as there are samples in the
bin.

The result in Equation \ref{XRef-Equation-41124426} can be summarised
as follows for $n\geq 1$ (note that the r.h.s. is 0 for $n=0$)
\begin{equation}
a_{i} {\left( {a_{j}}^{\dagger }\right) }^{n}\left| 0\right\rangle
=n \delta _{i,j} {\left( {a_{j}}^{\dagger }\right) }^{n-1}\left|
0\right\rangle  %
\label{XRef-Equation-41124843}
\end{equation}

\noindent which can be represented for a 4-bin histogram and for
$j=3$ as
\begin{equation}
\begin{array}{ccccc}
 \left( 0,0,0,0\right)  & \overset{{\left( {a_{j}}^{\dagger }\right)
}^{n}}{\longrightarrow } & \left( 0,0,n,0\right)  & \overset{a_{i}}{\longrightarrow
} & \begin{array}{cc}
 n \left( 0,0,n-1,0\right)  & i=j \\
 0 & i\neq j
\end{array}
\end{array}
\end{equation}

\noindent This result may be used in general to move annihilation
operators to the right of all creation operators. The result in
Equation \ref{XRef-Equation-41124843} is easily proved by using
$[a_{i},{a_{j}}^{\dagger }]=\delta _{i,j}$ to progressively move
$a_{i}$ to the right through one ${a_{j}}^{\dagger }$ at a time,
and then using $a_{i}|0\rangle =0$ to discard any terms that contain
$a_{i}|0\rangle $.
\subsubsection{Doing Calculations with Creation and Annihilation
Operators}

Using explicit notation (e.g. $(0,0,0,0)\overset{{a_{i}}^{\dagger
}}{\longrightarrow }(0,0,1,0)$) for what the creation and annihilation
operators are doing to the samples in the histogram bins is very
tedious in cases that are not much more complicated than the ones
discussed above. The purpose of introducing creation and annihilation
operators is to replace the manipulation of histogram samples by
algebraic manipulations based on the properties $a_{i}|0\rangle
=0$ and $[a_{i},{a_{j}}^{\dagger }]=\delta _{i,j}$, which also has
the desirable side effect that the calculations can be completely
automated by using symbolic algebra techniques. In general, explicit
notation\ \ should be needed only to verify what is being done to
the samples in the histograms, and to check that this corresponds
to what was intended.

From a theoretical point of view the commutation relations in Equation
\ref{XRef-Equation-32216132} are an algebraic way of doing the book-keeping
to keep track of how creation and annihilation operators construct
and modify histogram states depending on the order in which the
operators are applied. The $[a_{i},{a_{j}}^{\dagger }]=\delta _{i,j}$
commutation relation can be written in the form $a_{i}{a_{j}}^{\dagger
}={a_{j}}^{\dagger }a_{i}+\delta _{i,j}$, which can then used to
replace\ \ $a_{i}{a_{j}}^{\dagger }$ by ${a_{j}}^{\dagger }a_{i}+\delta
_{i,j}$, which effectively moves the annihilation operator to the
right (giving the ${a_{j}}^{\dagger }a_{i}$ term) whilst picking
up a commutator (the $\delta _{i,j}$ term) as a side effect. This
says that annihilation after creation (i.e. $a_{i}{a_{j}}^{\dagger
}$) is the same as annihilation before creation (i.e. ${a_{j}}^{\dagger
}a_{i}$), except for when the operators are applied to the same
bin, which triggers the appearance of the $\delta _{i,j}$ term for
reasons discussed above.

As a manual exercise, it can be verified that operators with the
above properties (i.e. $a_{i}|0\rangle =0$ and $[a_{i},{a_{j}}^{\dagger
}]=\delta _{i,j}$) correctly annihilate a sample from a 2-sample
histogram (samples in any bins); this generalises Equation \ref{XRef-Equation-33114267}
to the case where the bins are {\itshape not} assumed to be the
same. The strategy in this derivation (and in all other derivations
using creation and annihilation operators) is to move the annihilation
operators to the right of all the creation operators (using $a_{i}{a_{j}}^{\dagger
}={a_{j}}^{\dagger }a_{i}+\delta _{i,j}$), thus generating a sum
of terms of the form $(a^{\dagger }a^{\dagger }a^{\dagger }a^{\dagger
} \cdots )(a a a a \cdots )|0\rangle $, and wherever there is a
non-zero number of annihilation operators acting on $|0\rangle $
the term may be removed (using $a_{i}|0\rangle =0$). This leaves
a sum of terms that contain only creation operators acting on $|0\rangle
$.

The detailed derivation of the effect of applying $\sum _{j=1}^{m}a_{j}$
to ${a_{i_{1}}}^{\dagger }{a_{i_{2}}}^{\dagger }|0\rangle $ is as
follows
\begin{equation}
\begin{array}{rl}
 \left. \left( \sum _{j=1}^{m}a_{j}\right)  {a_{i_{1}}}^{\dagger
}{a_{i_{2}}}^{\dagger }\left| 0\right. \right\rangle   & =\left(
\sum _{j=1}^{m}a_{j}{a_{i_{1}}}^{\dagger }\right) {a_{i_{2}}}^{\dagger
} \left. \left| 0\right. \right\rangle   \\
  & =\sum _{j=1}^{m}\left( {a_{i_{1}}}^{\dagger }a_{j}+\delta _{i_{1},j}\right)
{a_{i_{2}}}^{\dagger } \left. \left| 0\right. \right\rangle   \\
  & =\sum _{j=1}^{m}\left( {a_{i_{1}}}^{\dagger }( a_{j}{a_{i_{2}}}^{\dagger
}) +\delta _{i_{1},j}{a_{i_{2}}}^{\dagger }\right)  \left. \left|
0\right. \right\rangle   \\
  & =\sum _{j=1}^{m}\left( {a_{i_{1}}}^{\dagger }( {a_{i_{2}}}^{\dagger
}a_{j}+\delta _{i_{2},j}) +\delta _{i_{1},j}{a_{i_{2}}}^{\dagger
}\right)  \left. \left| 0\right. \right\rangle   \\
  & =\sum _{j=1}^{m}\left( {a_{i_{1}}}^{\dagger }{a_{i_{2}}}^{\dagger
}a_{j}+\delta _{i_{2},j}{a_{i_{1}}}^{\dagger }+\delta _{i_{1},j}{a_{i_{2}}}^{\dagger
}\right)  \left. \left| 0\right. \right\rangle   \\
  & =\sum _{j=1}^{m}\left( {a_{i_{1}}}^{\dagger }{a_{i_{2}}}^{\dagger
}( \left. a_{j} \left| 0\right. \right\rangle  ) +\delta _{i_{2},j}(
{a_{i_{1}}}^{\dagger } \left. \left| 0\right. \right\rangle  ) +\delta
_{i_{1},j}( {a_{i_{2}}}^{\dagger } \left. \left| 0\right. \right\rangle
) \right)  \\
  & ={a_{i_{1}}}^{\dagger } \left. \left| 0\right. \right\rangle
+{a_{i_{2}}}^{\dagger } \left. \left| 0\right. \right\rangle
\end{array}%
\label{XRef-Equation-322162635}
\end{equation}

\noindent After this sort of manipulation has been done a few times
it is not necessary to write down all of the intermediate steps
as above, because the manipulations have a very simple form where
each annihilation operator $a_{i}$ is moved freely to the right,
except that whenever it passes through a corresponding creation
operator ${a_{j}}^{\dagger }$ an additional term is created (i.e.
the $\delta _{i,j}$ commutator term). In more complicated cases
it is more convenient to replace manual manipulations with symbolic
manipulations.
\subsubsection{Number Operator}

The above results (e.g. see Equation \ref{XRef-Equation-41124426})
allow the definition of a {\itshape number operator} $\mathcal{N}$
that counts the total number of samples in the histogram. Thus
\begin{equation}
\mathcal{N}\equiv \sum _{i=1}^{m}{a_{i}}^{\dagger }a_{i}%
\label{XRef-Equation-331144329}
\end{equation}

\noindent This gives
\begin{equation}
\mathcal{N} {\left( {a_{1}}^{\dagger }\right) }^{n_{1}}{\left( {a_{2}}^{\dagger
}\right) }^{n_{2}}\cdots  {\left( {a_{m}}^{\dagger }\right) }^{n_{m}}
\left| 0\right\rangle  =\left( n_{1}+n_{2}+\cdots  +n_{m}\right)
{\left( {a_{1}}^{\dagger }\right) }^{n_{1}}{\left( {a_{2}}^{\dagger
}\right) }^{n_{2}}\cdots  {\left( {a_{m}}^{\dagger }\right) }^{n_{m}}
\left| 0\right\rangle
\end{equation}

\noindent where the {\itshape total} number of histogram samples
$n\equiv n_{1}+n_{2}+\cdots  +n_{m}$ is the quantity that is measured
by applying $\mathcal{N}$. For instance, $\mathcal{N} {({a_{j}}^{\dagger
})}^{n_{j}}|0\rangle $ can be represented for a 4-bin histogram
and for $j=3$ as
\begin{equation}
\begin{array}{ccccc}
 \left( 0,0,0,0\right)  & \overset{{\left( {a_{j}}^{\dagger }\right)
}^{n_{j}}}{\longrightarrow } & \left( 0,0,n_{j},0\right)  & \overset{\mathcal{N}}{\longrightarrow
} & n_{j} \left( 0,0,n_{j},0\right)
\end{array}
\end{equation}

\noindent The structure of $\mathcal{N}$ in Equation \ref{XRef-Equation-331144329}
makes it clear how to define the number operator $\mathcal{N}_{i}$
for bin $i$ of the histogram, so that $\mathcal{N}=\sum _{i=1}^{m}\mathcal{N}_{i}$
where $\mathcal{N}_{i}$ is defined as
\begin{equation}
\mathcal{N}_{i}\equiv {a_{i}}^{\dagger }a_{i}
\end{equation}

\noindent and $\mathcal{N}_{i} {({a_{j}}^{\dagger })}^{n_{j}}|0\rangle
$ may be represented for a 4-bin histogram and for $j=3$ as
\begin{equation}
\begin{array}{ccccc}
 \left( 0,0,0,0\right)  & \overset{{\left( {a_{j}}^{\dagger }\right)
}^{n_{j}}}{\longrightarrow } & \left( 0,0,n_{j},0\right)  & \overset{\mathcal{N}_{j}}{\longrightarrow
} & \begin{array}{cc}
 n_{j} \left( 0,0,n_{j},0\right)  & i=j \\
 0 & i\neq j
\end{array}
\end{array}
\end{equation}
\subsubsection{Orthogonality and Completeness}\label{XRef-Subsubsection-11714376}

The states constructed using the creation operators described above
are orthogonal and complete. Consider the general histogram state
${({a_{1}}^{\dagger })}^{n_{1}}{({a_{2}}^{\dagger })}^{n_{2}}\cdots
{({a_{m}}^{\dagger })}^{n_{m}} |0\rangle $ and attempt to annihilate
its samples. The strategy of the proof will be to demonstrate that
there is a \textit{unique} set of annihilation operators that you
have to use in order to recover the empty histogram state $|0\rangle
$.

Apply a single annihilation operator $a_{1}$ (using Equation \ref{XRef-Equation-41124843}
to move it to the right)
\begin{equation}
a_{1} {\left( {a_{1}}^{\dagger }\right) }^{n_{1}}{\left( {a_{2}}^{\dagger
}\right) }^{n_{2}}\cdots  {\left( {a_{m}}^{\dagger }\right) }^{n_{m}}
\left| 0\right\rangle  ={n_{1}( {a_{1}}^{\dagger }) }^{n_{1}-1}{\left(
{a_{2}}^{\dagger }\right) }^{n_{2}}\cdots  {\left( {a_{m}}^{\dagger
}\right) }^{n_{m}} \left| 0\right\rangle
\end{equation}

\noindent Now apply the same annihilation operator $n_{1}-1$ more
times to eventually obtain
\begin{equation}
{\left( a_{1}\right) }^{n_{1}}{\left( {a_{1}}^{\dagger }\right)
}^{n_{1}}{\left( {a_{2}}^{\dagger }\right) }^{n_{2}}\cdots  {\left(
{a_{m}}^{\dagger }\right) }^{n_{m}} \left| 0\right\rangle  =n_{1}!{\left(
{a_{2}}^{\dagger }\right) }^{n_{2}}\cdots  {\left( {a_{m}}^{\dagger
}\right) }^{n_{m}} \left| 0\right\rangle
\end{equation}

\noindent Repeat this pattern of annihilation successively for bins
$2,3,\cdots ,m$ of the histogram to obtain
\begin{equation}
{\left( a_{m}\right) }^{n_{m}}\cdots  {\left( a_{2}\right) }^{n_{2}}{\left(
a_{1}\right) }^{n_{1}}{\left( {a_{1}}^{\dagger }\right) }^{n_{1}}{\left(
{a_{2}}^{\dagger }\right) }^{n_{2}}\cdots  {\left( {a_{m}}^{\dagger
}\right) }^{n_{m}} \left| 0\right\rangle  =n_{1}!n_{2}!\cdots  n_{m}!
\left| 0\right\rangle  %
\label{XRef-Equation-513124511}
\end{equation}

\noindent where the resulting state is (proportional to) the empty
histogram $|0\rangle $.

Thus we recover the empty histogram by applying exactly those annihilation
operators to the histogram that correspond to the creation operators
that we used to construct the histogram in the first place. The
fact that the empty histogram can be recovered {\itshape only} by
applying the {\itshape same} set $(n_{1},n_{2},\cdots  ,n_{m})$
of annihilation operators as creation operators means that the states
are {\itshape orthogonal}, and the fact that {\itshape all} possible
states are constructable using the appropriate set $(n_{1},n_{2},\cdots
,n_{m})$ of creation operators means that the states are {\itshape
complete}.

The constant of proportionality $n_{1}!n_{2}!\cdots  n_{m}!$ is
the number of ways in which the annihilation operators can annihilate
the histogram samples, which corresponds to the total number of
ways of permuting the samples within the histogram bins (but {\itshape
not} permuting between bins). If this permutation factor is not
required then the states could be defined as $\frac{1}{\sqrt{n_{1}!n_{2}!\cdots
n_{m}!}}{({a_{1}}^{\dagger })}^{n_{1}}{({a_{2}}^{\dagger })}^{n_{2}}\cdots
{({a_{m}}^{\dagger })}^{n_{m}} |0\rangle $, and a similar normalisation
factor $\frac{1}{\sqrt{n_{1}!n_{2}!\cdots  n_{m}!}}$ should be included
with the annihilation operators when this whole state is to be annihilated.
It is a matter of tast whether the normalisation factor \textit{is}
included along with the state, or whether it is \textit{not} included
but is then subsequently divided out from the results of calculations.
\subsubsection{States and Adjoint States}\label{XRef-Subsubsection-116143013}

The above results on orthogonality and completeness can be written
more rigorously by introducing the {\itshape adjoint} state. Intuitively,
the adjoint state is obtained by time-reversing everything, so that
instead of making operators act to the right (with operators that
act later being placed further to the left), the operators in an
adjoint state act to the left (with operators that act earlier being
placed further to the right). Note that between these two viewpoints
the time order of operator action corresponds to the order in which
the operators appear in the ``operator product''. Also note that
a creation operator acting to the right (i.e. create a sample as
time increases, as in ${a_{i}}^{\dagger }|0\rangle $) behaves in
the same way as an annihilation operator acting to the left (i.e.
annihilate a sample as time decreases, as in $\langle 0|{a_{i}}^{\dagger
}=0$). In this case ${a_{i}}^{\dagger }|0\rangle $ says (reading
from right to left) that there is an empty histogram in the distant
past which later has a sample created in bin $i$, whereas $\langle
0|{a_{i}}^{\dagger }$ says (reading from left to right) that there
is an empty histogram in the distant future which earlier has a
sample annihilated from bin $i$ to give 0 (i.e. $\langle 0|{a_{i}}^{\dagger
}=0$).

Introduce a notation for a histogram with occupancies $(n_{1},n_{2},\cdots
,n_{m})$
\begin{equation}
\Theta _{n_{1},n_{2},\cdots ,n_{m}}\equiv {\left( {a_{1}}^{\dagger
}\right) }^{n_{1}}{\left( {a_{2}}^{\dagger }\right) }^{n_{2}}\cdots
{\left( {a_{m}}^{\dagger }\right) }^{n_{m}} \left| 0\right\rangle
\label{XRef-Equation-11715150}
\end{equation}

\noindent and its adjoint state for creating a histogram with occupancies
$(n_{1},n_{2},\cdots ,n_{m})$, {\itshape but} done in the reversed
time sense where there is an empty histogram in the far future,
which is then populated as we move backwards in time
\begin{equation}
{\Theta ^{\dagger }}_{n_{1},n_{2},\cdots ,n_{m}}=\left\langle  0\right|
{\left( a_{m}\right) }^{n_{m}}\cdots  {\left( a_{2}\right) }^{n_{2}}{\left(
a_{1}\right) }^{n_{1}}%
\label{XRef-Equation-520162357}
\end{equation}

\noindent The orthogonality property can then be stated as
\begin{equation}
{\Theta ^{\dagger }}_{\nu _{1},\nu _{2},\cdots ,\nu _{m}}\Theta
_{n_{1},n_{2},\cdots ,n_{m}}=\delta _{n_{1},\nu _{1}}\delta _{n_{2},\nu
_{2}}\cdots  \delta _{n_{m},\nu _{m}}n_{1}!n_{2}!\cdots  n_{m}!
\end{equation}

\noindent where the result in Equation \ref{XRef-Equation-513124511}
is used, and where $\langle 0||0\rangle \equiv 1$ is defined. The
completeness property then corresponds to the following resolution
of the identity operator
\begin{equation}
\sum _{n_{1},n_{2},\cdots ,n_{m}}\frac{1}{n_{1}!n_{2}!\cdots  n_{m}!}\Theta
_{n_{1},n_{2},\cdots ,n_{m}}{\Theta ^{\dagger }}_{n_{1},n_{2},\cdots
,n_{m}}=1
\end{equation}

\noindent where the states that this operator acts upon are assumed
to be constructed in the same way as $\Theta _{n_{1},n_{2},\cdots
,n_{m}}$ (i.e. using creation operators).
\subsubsection{Summary of Useful Results}\label{XRef-Subsubsection-117133527}
\begin{enumerate}
\item Creation operator for bin $i$: ${a_{i}}^{\dagger }$. When
applied to a histogram state this creates one sample in bin $i$.
\item Annihilation operator for bin $i$: $a_{i}$. When applied to
a histogram state this annihilates one sample from bin $i$ in as
many ways (i.e. $n_{i}$) as there are samples already in bin $i$.
The result is $n_{i}$ copies of the histogram state with one sample
annihilated from bin $i$. This includes the special case $n_{i}=0$
where the histogram is annihilated altogether to give $0$.
\item Annihilation operator for all bins: $\sum _{i=1}^{m}a_{i}$.
This produces a generalisation of what $a_{i}$ alone does. For each
$i$ ($i=1,2,\cdots ,m$) the result is $n_{i}$ copies of the histogram
state with one sample annihilated from bin $i$, which gives a total
of $\sum _{i=1}^{m}n_{i}$ histograms. This operator is useful for
preparing a histogram for an MCMC update because it removes a sample
at random from the histogram (i.e. it prepares $\sum _{i=1}^{m}n_{i}$
copies of the histogram in each of which a different sample has
been annihilated).
\item Annihilate an empty histogram: $a_{i}|0\rangle =0$. This defines
the ``vacuum'' state as a reference state for determining the occupancy
of each histogram bin. This definition is very useful for removing
terms that do not contribute to the overall histogram state.
\item Creation/annihilation commutator: $[a_{i},{a_{j}}^{\dagger
}]=\delta _{i,j}$. This summarises the basic interaction between
the creation and annihilation operators. It is mainly used in the
form $a_{i}{a_{j}}^{\dagger }={a_{j}}^{\dagger }a_{i}+\delta _{i,j}$
to move annihilation operators to the right of creation operators,
which eventually brings the annihilation operators so that they
act directly on $|0\rangle $, where they can be removed (using $a_{i}|0\rangle
=0$).
\item Annihilation/annihilation and creation/creation commutators:
$[a_{i},a_{j}]=0$ and $[{a_{i}}^{\dagger },{a_{j}}^{\dagger }]=0$.
These summarise the fact that a sequence consisting solely of annihilation
operations (or solely of creation operations) has the same effect
whatever the order in which the operators appear in the sequence.
\item Moving an annihilation operator to the right: $a_{i} {({a_{j}}^{\dagger
})}^{n}|0\rangle =n \delta _{i,j} {({a_{j}}^{\dagger })}^{n-1}|0\rangle
$: This is the basic result that is used to remove annihilation
operators from expressions. The $a_{i}$ is moved progressively to
the right through the ${a_{j}}^{\dagger }$ (using $a_{i}{a_{j}}^{\dagger
}={a_{j}}^{\dagger }a_{i}+\delta _{i,j}$) until it reaches the $|0\rangle
$, where it is discarded (using $a_{i}|0\rangle =0$).
\item Number operator for bin $i$: $\mathcal{N}_{i}={a_{i}}^{\dagger
}a_{i}$. This annihilates then creates a sample in bin $i$. Because
there are $n_{i}$ ways of annihilating a sample but only 1 way of
creating a sample, the net effect is to count the number $n_{i}$
of samples in bin $i$.
\item Total number operator for all bins: $\mathcal{N}=\sum _{i=1}^{m}{a_{i}}^{\dagger
}a_{i}$. This counts the total number of samples in the histogram.
This follows directly from $\mathcal{N}_{i}={a_{i}}^{\dagger }a_{i}$
above.
\item State and adjoint state: $\Theta _{n_{1},n_{2},\cdots ,n_{m}}={({a_{1}}^{\dagger
})}^{n_{1}}{({a_{2}}^{\dagger })}^{n_{2}}\cdots  {({a_{m}}^{\dagger
})}^{n_{m}} |0\rangle $ and ${\Theta ^{\dagger }}_{n_{1},n_{2},\cdots
,n_{m}}=\langle 0|{(a_{m})}^{n_{m}}\cdots  {(a_{2})}^{n_{2}}{(a_{1})}^{n_{1}}$
(respectively). The adjoint state can be applied to the left of
a state and the annihilation operators then moved to the right using
$a_{i} {({a_{j}}^{\dagger })}^{n}|0\rangle =n \delta _{i,j} {({a_{j}}^{\dagger
})}^{n-1}|0\rangle $ to demonstrate orthogonality (assuming $\langle
0||0\rangle \equiv 1$). The adjoint of $a_{i}|0\rangle =0$ implies
$\langle 0|{a_{i}}^{\dagger }=0$.
\item Orthogonality:\ \ ${\Theta ^{\dagger }}_{\nu _{1},\nu _{2},\cdots
,\nu _{m}}\Theta _{n_{1},n_{2},\cdots ,n_{m}}=\delta _{n_{1},\nu
_{1}}\delta _{n_{2},\nu _{2}}\cdots  \delta _{n_{m},\nu _{m}}n_{1}!n_{2}!\cdots
n_{m}!$. Here $\langle 0||0\rangle \equiv 1$ is assumed by definition.
\item Completeness: All states $\Theta _{n_{1},n_{2},\cdots ,n_{m}}$
are constructable by using the appropriate set $(n_{1},n_{2},\cdots
,n_{m})$ of creation operators.
\end{enumerate}
\subsubsection{Multiple MRF Nodes}

The above results are for a \textit{single} MRF node. When there
are multiple nodes, each MRF node has it own set of creation and
annihilation operators, which have all of the properties described
above. Operators for different nodes commute with each other because
they act on different state spaces, so the generalised form of Equation
\ref{XRef-Equation-32216132} is
\begin{equation}
\begin{array}{rl}
 \left[ a_{i}^{s},a_{j}^{t \dagger }\right]  & =\delta _{i,j}\delta
_{s,t} \\
 \left[ a_{i}^{s},a_{j}^{t}\right]  & =0 \\
 \left[ a_{i}^{s \dagger },a_{j}^{t \dagger }\right]  & =0
\end{array}
\end{equation}

\noindent where $s$ and $t$ are node indices. There are analogous
generalisations of all the results in Section \ref{XRef-Subsubsection-117133527}.
\subsection{MCMC Update Operator}\label{XRef-Subsection-322174917}

In Section \ref{XRef-Subsubsection-331134331} it was shown how the
state of an $N$-node MRF can be represented as a set of $N$ histograms
each of which contains one sample in one of the histogram bins,
and how MCMC updates of the MRF can be represented as hopping operations
where each sample hops around between the bins of its histogram.
The aim now is to use the creation and annihilation operators defined
in Section \ref{XRef-Subsubsection-331145022} to implement these
MCMC hopping operations.

The MCMC update operator $\mathcal{H}$ can be constructed in several
easy steps, in which each MCMC hopping operation is broken down
into annihilation followed by subsequent creation of a sample.
\begin{enumerate}
\item Annihilate a sample (see the middle row of Figure \ref{XRef-FigureCaption-101810312}).
Apply $\sum _{j=1}^{m}a_{j}$ to the histogram state to annihilate
one sample from each bin, which prepares $\sum _{i=1}^{m}n_{i}$
copies of the histogram in each of which a different sample has
been annihilated. The output of this operation is thus a linear
combination of histogram states, where each state is weighted by
the same factor of unity (i.e. all states are equally likely). This
linear combination of $\sum _{i=1}^{m}n_{i}$ terms (of which only
$m$ are distinct) represents the ensemble of all the possible outcomes
of annihilating one sample.
\item Create a sample (see the bottom row of Figure \ref{XRef-FigureCaption-101810312}).
Apply $\sum _{i=1}^{m}p_{i}{a_{i}}^{\dagger }$ to each histogram
state in the ensemble generated above, which prepares $m$ copies
of the histogram in each of which a different sample has been created,
and weight each of these $m$ histogram states so that where the
sample is created in bin $i$ the state is weighted by a factor $p_{i}$.
If the $p_{i}$ satisfy $p_{i}\geq 0$ and $\sum _{i=1}^{m}p_{i}=1$
then $p_{i}$ can be interpreted as the probability of creating a
sample in bin $i$. Actually, the normalisation condition $\sum _{i=1}^{m}p_{i}=1$
can be omitted because the {\itshape relative }size of the $p_{i}$
is all that is required. The output of this operation is thus a
linear combination of histogram states, where each state is weighted
by the appropriate probability factor $p_{i}$ corresponding to the
bin $i$ in which a sample has just been created. This linear combination
of $m$ terms represents the ensemble of all the possible outcomes
of creating one sample in one of the bins of a histogram.
\end{enumerate}

\noindent Concatenate these two operators to define the MCMC update
operator $\mathcal{H}$
\begin{equation}
\mathcal{H}\equiv \sum _{i=1}^{m}p_{i}{a_{i}}^{\dagger }\sum _{j=1}^{m}a_{j}%
\label{XRef-Equation-1018122141}
\end{equation}

\noindent where the action of $\sum _{j=1}^{m}a_{j}$ produces $\sum
_{i=1}^{m}n_{i}$ histograms, then the action of $\sum _{i=1}^{m}p_{i}{a_{i}}^{\dagger
}$ on \textit{each} of these $\sum _{i=1}^{m}n_{i}$ histograms produces
$m$ histograms. Finally, all of these histograms should be regrouped
so that multiple copies of identical histograms are represented
as a single copy with an appropriate weighting factor.

The weighting factor that is applied to the state (as used here)
represents probability itself rather than probability amplitude
(as used in the corresponding QFT). However, if a QFT is ``Wick
rotated'' to become a Euclidean QFT then it is equivalent to quantum
statistical mechanics \cite{Zee2003}, where the state is a probability-weighted
mixture of states. So the\ \ approach discussed in this paper has
a mathematical structure that is similar to the Euclidean version
of a QFT of bosons.

The pieces $p_{i}{a_{i}}^{\dagger }a_{j}$ of the MCMC update operator
may be represented diagrammatically as
\[
p_{i} \left( \begin{array}{ccccc}
 j & \overset{a_{j}}{\longrightarrow } & \cdot  & \overset{{a_{i}}^{\dagger
}}{\longrightarrow } & i \\
   &   & \Uparrow  &   &   \\
   &   & \mathrm{source} &   &
\end{array}\right)
\]

\noindent where state $j$ comes in from the left and is annihilated
by $a_{j}$, and a new state $i$ is created by ${a_{i}}^{\dagger
}$ which then goes out to the right, and the probability of this
transition occurring is $p_{i}$ which depends only on the output
state (so it is memoryless), which is in turn generated by a source
(e.g. MRF neighbours, external source, etc). The whole MCMC update
operator $\mathcal{H}$ is the sum of this diagram over states $i$
and $j$.

This result can be generalised to an MRF with $N$ nodes (with node
$s$ having $m_{s}$ states)
\begin{equation}
\mathcal{H}\longrightarrow \sum _{s=1}^{N}\sum _{i=1}^{m_{s}}p_{i}^{s}
a_{i}^{s \dagger }\sum _{j=1}^{m_{s}}a_{j}^{s}%
\label{XRef-Equation-51716233}
\end{equation}

\noindent which can be written using the transition operator $\mathcal{T}_{i,j}^{s}\equiv
a_{i}^{s \dagger }a_{j}^{s}$ that hops a sample from bin $j$ to
bin $i$ at node $s$.
\begin{equation}
\mathcal{H}=\sum _{s=1}^{N}\sum _{i,j=1}^{m_{s}}p_{i}^{s} \mathcal{T}_{i,j}^{s}%
\label{XRef-Equation-41103321}
\end{equation}

In practice the creation probability $p_{i}^{s}$ depends (via a
product of clique factors, as described in the discussion on the
HCE in Section \ref{XRef-Subsubsection-331132456}) on the states
of the other nodes in the MRF. This probability can be computed
by applying an appropriately designed operator to the MRF node states.
Thus use the number operator for bin $k$ at node $t$ (which is $\mathcal{N}_{k}^{t}\equiv
a_{k}^{t \dagger }a_{k}^{t}$) weighted by $p_{i,k}^{s,t}$ to determine
the 2-clique contribution (i.e. pairwise interactions between nodes
of the MRF) for creation in bin $i$ at node $s$ due to bin $k$ at
node $t$ being occupied. This operator expression is appropriate
for {\itshape any} number of samples in bin $k$ at node $t$, because
the number operator $\mathcal{N}_{k}^{t}$ automatically determines
the number of samples as needed, and then uses this number to weight
any clique factor that involves this node.

This use of sample number to weight clique factors is consistent
because it guarantees that a {\itshape single} sample at each node
(i.e. standard HCE) is physically equivalent to the situation where
each of these samples is cut into a number of equal-sized sub-samples,
because the additional factors then generated by the number operator
applied to these sub-samples are exactly cancelled by the additional
factors then generated by the fact that interactions between {\itshape
sub}-samples are proportionally weaker than interactions between
samples.

This allows $p_{i}^{s}$ to be replaced by an operator $\mathcal{P}_{i}^{s}$,
which can be used to construct a $p_{i}^{s}$ based on whatever samples
it finds in the histograms in the neighbourhood $C( s) $ of node
$s$ of the MRF.
\begin{equation}
p_{i}^{s}\longrightarrow \mathcal{P}_{i}^{s}\equiv \prod _{t\in
C( s) }\sum _{k=1}^{m_{t}}p_{i,k}^{s,t}\mathcal{N}_{k}^{t}%
\label{XRef-Equation-52015372}
\end{equation}

\noindent This result should be compared with the product form of
the HCE in Equation \ref{XRef-Equation-322152418}, where the $\prod
_{t\in C( s) }(\cdots )$ in Equation \ref{XRef-Equation-52015372}
corresponds to the $\prod _{c}(\cdots )$ in Equation \ref{XRef-Equation-322152418},
and the sum over operators $\sum _{k=1}^{m_{t}}(\cdots )$ in Equation
\ref{XRef-Equation-52015372}\ \ is needed to cover all the possibilities
that might appear in the $(\cdots )$ inside $\prod _{c}(\cdots )$
in Equation \ref{XRef-Equation-322152418}. More generally for 3-cliques
the operator $\mathcal{P}_{i}^{s}$ is given by
\begin{equation}
p_{i}^{s}\longrightarrow \mathcal{P}_{i}^{s}\equiv \prod _{t_{1},t_{2}\in
C( s) }\sum _{k_{1}=1}^{m_{t_{1}}}\sum _{k_{2}=1}^{m_{t_{2}}}p_{i,k_{1},k_{2}}^{s,t_{1},t_{2}}\mathcal{N}_{k_{1}}^{t_{1}}\mathcal{N}_{k_{2}}^{t_{2}}%
\label{XRef-Equation-1018131316}
\end{equation}

\noindent which may be straightforwardly generalised to higher order
cliques.

Inserting the operator-valued version of $p_{i}^{s}$ into Equation
\ref{XRef-Equation-41103321}, the MCMC update operator $\mathcal{H}$
becomes (using 2-cliques only)
\begin{equation}
\mathcal{H}\longrightarrow \sum _{s=1}^{N}\sum _{i,j=1}^{m_{s}}\mathcal{T}_{i,j}^{s}
\left( \prod _{t\in C( s) }\sum _{k=1}^{m_{t}}p_{i,k}^{s,t}\mathcal{N}_{k}^{t}\right)
\label{XRef-Equation-41104529}
\end{equation}

\noindent with analogous expressions for higher order cliques. This
operator-valued object $\mathcal{H}$ can be applied to {\itshape
any} MRF state, whether it is a conventional {\itshape single} sample
per node state, or has {\itshape multiple} samples per node. This
is the key advantage of using operators, because they are effectively
general procedures (e.g. algorithms) that can be applied to any
state that is constructed using creation operators. The algebra
of the creation and annihilation operators provides a unified framework
for handing all of these possibilities consistently.

The functional form used in Equation \ref{XRef-Equation-41104529}
is enforced by backward compatibility with the MCMC update operator
for an MRF shown in Equation \ref{XRef-Equation-51716233}, where
the factor $p_{i}^{s}$ is a product of clique factors that intersect
with node $s$ (i.e. for 2-cliques only, it is generated by the $\prod
_{t\in C( s) }\sum _{k=1}^{m_{t}}p_{i,k}^{s,t}\mathcal{N}_{k}^{t}$
factor in Equation \ref{XRef-Equation-41104529}). However, the framework
developed here allows for any functional form built out of creation
and annihilation operators, so a very large class of update operators
$\mathcal{H}$ can be constructed such as:
\begin{enumerate}
\item The operator that generates the product of clique factors
$\prod _{t\in C( s) }\sum _{k=1}^{m_{t}}p_{i,k}^{s,t}\mathcal{N}_{k}^{t}$
can be replaced by some other functional form, such as a non-linear
sigmoid squashing function $\sigma ( \sum _{t\in C( s) }\sum _{k=1}^{m_{t}}p_{i,k}^{s,t}\mathcal{N}_{k}^{t})
$, as is typically done in ``neural network'' implementations of
recurrent networks. One possible way of viewing the relationship
between this non-linear sigmoidal version and the clique product
can be obtained by perturbatively expanding the sigmoid to obtain
various powers of its argument $\sum _{t\in C( s) }\sum _{k=1}^{m_{t}}p_{i,k}^{s,t}\mathcal{N}_{k}^{t}$,
which includes terms that look like the original clique product
$\prod _{t\in C( s) }\sum _{k=1}^{m_{t}}p_{i,k}^{s,t}\mathcal{N}_{k}^{t}$,
plus other higher order terms.
\item The hopping operator $\mathcal{T}_{i,j}^{s}=a_{i}^{s \dagger
}a_{j}^{s}$ can be replaced by some other functional form, such
as one that increases (i.e. birth) or decreases (i.e. death) the
number of samples, which may be used to allow the update operator
$\mathcal{H}$ to explore histogram states with various occupancies.
Note that if this part of the overall update operator $\mathcal{H}$
is used \textit{alone} as the update operator (i.e. without the
clique factor piece above), then it can be used to generate the
prior behaviour that the histogram state has before any interactions
with other histograms are included.
\end{enumerate}

\noindent The effect of the creation and annihilation operators
can be viewed in terms of elementary operations on histograms (as
described in Section \ref{XRef-Subsubsection-331145022}), and their
operator algebra can be used to do calculations in which $\mathcal{H}$
is applied to multiply occupied states to generate MCMC updates.
It is also possible to use symbolic algebra to do these operator
manipulations automatically. In general, the effect of the MCMC
update operator $\mathcal{H}$ on a set of histogram states can be
represented as a type of Feynman diagram, in which each vertex represents
a product of operators acting on an incoming state to produce an
outgoing state (if any), and a (weighted) sum of such diagrams represents
the corresponding (weighted) sum of products of operators (note
that here the weights are probabilities rather than probability
amplitudes).

Note that the MCMC update operator $\mathcal{H}$ in Equation \ref{XRef-Equation-41104529}
is number-conserving in the sense that its transition operator $\mathcal{T}_{i,j}^{s}\equiv
a_{i}^{s \dagger }a_{j}^{s}$ causes samples to hop from bin $j$
to bin $i$ at node $s$, without gain or loss of the total number
of samples at node $s$. Formally, this property may be written as
$[\mathcal{H},\mathcal{N}^{s}]=0$ where $\mathcal{N}^{s}\equiv \sum
_{i=1}^{m_{s}}\mathcal{N}_{i}^{s}$ is the {\itshape total} number
operator at node $s$. This result can be seen intuitively because
it may be written as $\mathcal{H} \mathcal{N}^{s}=\mathcal{N}^{s}\mathcal{H}$,
which states that when you measure the total number of samples at
node $s$ then do an MCMC update, you get the \textit{same} result
as when you do an MCMC update then measure the total number of samples
at node $s$, so there must be number conservation.

The steps in the derivation of the number conservation property
$[\mathcal{H},\mathcal{N}^{u}]=0$ are as follows
\begin{equation}
\begin{array}{rl}
 \left[ \mathcal{H},\mathcal{N}^{u}\right]  & =\sum _{s=1}^{N}\sum
_{i,j=1}^{m_{s}}\left[ \mathcal{T}_{i,j}^{s} \left( \prod _{t\in
C( s) }\sum _{k=1}^{m_{t}}p_{i,k}^{s,t}\mathcal{N}_{k}^{t}\right)
,\mathcal{N}^{u}\right]  \\
  & =\sum _{s=1}^{N}\sum _{i,j=1}^{m_{s}}\left( \begin{array}{c}
 \mathcal{T}_{i,j}^{s} \left( \prod _{t\in C( s) }\sum _{k=1}^{m_{t}}p_{i,k}^{s,t}\mathcal{N}_{k}^{t}\right)
\mathcal{N}^{u} \\
 -\mathcal{N}^{u}\mathcal{T}_{i,j}^{s} \left( \prod _{t\in C( s)
}\sum _{k=1}^{m_{t}}p_{i,k}^{s,t}\mathcal{N}_{k}^{t}\right)
\end{array}\right)  \\
  & =\sum _{s=1}^{N}\sum _{i,j=1}^{m_{s}}\left( \begin{array}{c}
 \mathcal{T}_{i,j}^{s} \left( \prod _{t\in C( s) }\sum _{k=1}^{m_{t}}p_{i,k}^{s,t}\mathcal{N}_{k}^{t}\right)
\mathcal{N}^{u} \\
 -\mathcal{T}_{i,j}^{s} \mathcal{N}^{u}( \prod _{t\in C( s) }\sum
_{k=1}^{m_{t}}p_{i,k}^{s,t}\mathcal{N}_{k}^{t})
\end{array}\right)  \\
  & =\sum _{s=1}^{N}\sum _{i,j=1}^{m_{s}}\left( \begin{array}{c}
 \mathcal{T}_{i,j}^{s} \left( \prod _{t\in C( s) }\sum _{k=1}^{m_{t}}p_{i,k}^{s,t}\mathcal{N}_{k}^{t}\right)
\mathcal{N}^{u} \\
 -\mathcal{T}_{i,j}^{s} \left( \prod _{t\in C( s) }\sum _{k=1}^{m_{t}}p_{i,k}^{s,t}\mathcal{N}_{k}^{t}\right)
\mathcal{N}^{u}
\end{array}\right)  \\
  & =0
\end{array}
\end{equation}

\noindent using $[\mathcal{N}^{u},\mathcal{T}_{i,j}^{s}]=0$ ($\mathcal{T}_{i,j}^{s}$
causes hopping at node $s$ but conserves total number at node $s$,
\textit{and} also trivially conserves total number at all other
nodes) to make the replacement $\mathcal{N}^{u}\mathcal{T}_{i,j}^{s}\longrightarrow
\mathcal{T}_{i,j}^{s} \mathcal{N}^{u}$, and $[\mathcal{N}^{u},\mathcal{N}_{k}^{t}]=0$
(number operators always commute) to make the replacement $\mathcal{N}^{u}(
\prod _{t\in C( s) }\sum _{k=1}^{m_{t}}p_{i,k}^{s,t}\mathcal{N}_{k}^{t})
\longrightarrow (\prod _{t\in C( s) }\sum _{k=1}^{m_{t}}p_{i,k}^{s,t}\mathcal{N}_{k}^{t})\mathcal{N}^{u}$.
Note that the fact that $[\mathcal{N}^{u},\mathcal{T}_{i,j}^{s}]=0$
and $[\mathcal{N}^{u},\mathcal{N}_{k}^{t}]=0$ are simple to derive
from the basic creation/annihilation content of the various operators.

The overall effect of using creation and annihilation operators
is to formalise the act of manipulating samples in histograms, so
that these manipulations are now represented algebraically. One
\textit{could} avoid the use of this algebraic approach (especially
when each histogram has only a single sample, as in a standard MRF),
but as the manipulations become more complicated (e.g. subtle interdependencies
between histograms) it is better to do them by using this algebraic
approach.
\subsection{Diagrammatic Representation of MCMC Algorithms}\label{XRef-Subsubsection-91313212}

A sequence of MCMC updates (e.g. see Section \ref{XRef-Subsubsection-4611524})
in which $\textbf{\textit{x}}$ and $\textbf{\textit{y}}$ are \textit{alternately}
updated by sampling from $\Pr ( \textbf{\textit{x}},\textbf{\textit{y}})
$ is illustrated below where each arrow represents a dependency.
The graph structure shows that the updates are memoryless. For instance,
${\textbf{\textit{x}}}_{2}$ depends on ${\textbf{\textit{y}}}_{1}$
via $\Pr ( {\textbf{\textit{x}}}_{2}|{\textbf{\textit{y}}}_{1})
$, but it does \textit{not} depend on ${\textbf{\textit{x}}}_{1}$.

\[
\begin{array}{cccccccccc}
 {\textbf{\textit{x}}}_{1} &   & {\textbf{\textit{x}}}_{2} & \longrightarrow
& {\textbf{\textit{x}}}_{2} &   & {\textbf{\textit{x}}}_{3} & \longrightarrow
& {\textbf{\textit{x}}}_{3} & \cdots  \\
   & {}\begin{array}{ccc}
   &   & \nearrow  \\
   & \nearrow  &   \\
 \nearrow  &   &
\end{array} &   & \begin{array}{ccc}
 \searrow  &   &   \\
   & \searrow  &   \\
   &   & \searrow
\end{array} &   & {}\begin{array}{ccc}
   &   & \nearrow  \\
   & \nearrow  &   \\
 \nearrow  &   &
\end{array} &   & \begin{array}{ccc}
 \searrow  &   &   \\
   & \searrow  &   \\
   &   & \searrow
\end{array} &   & \cdots  \\
 {\textbf{\textit{y}}}_{1} & \longrightarrow  & {\textbf{\textit{y}}}_{1}
&   & {\textbf{\textit{y}}}_{2} & \longrightarrow  & {\textbf{\textit{y}}}_{2}
&   & {\textbf{\textit{y}}}_{3} & \cdots  \\
   &   &   &   &   &   &   &   &   &   \\
   & \Pr ( {\textbf{\textit{x}}}_{2}|{\textbf{\textit{y}}}_{1})
&   & \Pr ( {\textbf{\textit{y}}}_{2}|{\textbf{\textit{x}}}_{2})
&   & \Pr ( {\textbf{\textit{x}}}_{3}|{\textbf{\textit{y}}}_{2})
&   & \Pr ( {\textbf{\textit{y}}}_{3}|{\textbf{\textit{x}}}_{3})
&   &
\end{array}
\]

\noindent The above diagram can be skeletonised by omitting all
inessential labelling in order to emphasis the information flow,
in which case the result looks like this

\[
\begin{array}{cccccccccc}
 \cdot  &   & \cdot  & \longrightarrow  & \cdot  &   & \cdot  &
\longrightarrow  & \cdot  & \cdots  \\
   & {}\nearrow  &   & \searrow  &   & {}\nearrow  &   & \searrow
&   & \cdots  \\
 \cdot  & \longrightarrow  & \cdot  &   & \cdot  & \longrightarrow
& \cdot  &   & \cdot  & \cdots
\end{array}
\]

\noindent If this skeletonisation is used to draw an information
flow diagram for a sequence of MCMC updates of a 4 node Markov chain,
then a typical result looks like the diagram below.

\[
\begin{array}{ccccccccccccccccccccccc}
   & \cdot  & \longrightarrow  & \cdot  & \longrightarrow  & \cdot
& \longrightarrow  & \cdot  &   & \cdot  & \longrightarrow  & \cdot
& \longrightarrow  & \cdot  & \longrightarrow  & \cdot  & \longrightarrow
& \cdot  & \longrightarrow  & \cdot  & \longrightarrow  & \cdot
& \cdots  \\
 \pm 1 &   & \searrow  &   &   &   &   &   & \nearrow  &   &   &
&   &   & \searrow  &   &   &   &   &   & \ \  &   &   \\
   & \cdot  &   & \cdot  & \longrightarrow  & \cdot  & \longrightarrow
& \cdot  & \longrightarrow  & \cdot  &   & \cdot  &   & \cdot  &
& \cdot  & \longrightarrow  & \cdot  & \longrightarrow  & \cdot
& \longrightarrow  & \cdot  & \cdots  \\
 \pm 2 &   &   &   & \searrow  &   &   &   &   &   & \nearrow  &
& \nearrow  &   &   &   &   &   & \searrow  &   &   &   &   \\
   & \cdot  & \longrightarrow  & \cdot  &   & \cdot  &   & \cdot
& \longrightarrow  & \cdot  & \longrightarrow  & \cdot  & \longrightarrow
& \cdot  & \longrightarrow  & \cdot  &   & \cdot  &   & \cdot  &
& \cdot  & \cdots  \\
 \pm 3 &   &   &   &   &   & \nearrow  &   &   &   &   &   &   &
&   &   & \nearrow  &   &   &   & \nearrow  &   &   \\
   & \cdot  & \longrightarrow  & \cdot  & \longrightarrow  & \cdot
& \longrightarrow  & \cdot  & \longrightarrow  & \cdot  & \longrightarrow
& \cdot  & \longrightarrow  & \cdot  & \longrightarrow  & \cdot
& \longrightarrow  & \cdot  & \longrightarrow  & \cdot  & \longrightarrow
& \cdot  & \cdots  \\
   &   & +1 &   & +2 &   & -3 &   & -1 &   & -2 &   & -2 &   & +1
&   & -3 &   & +2 &   & -3 &   &
\end{array}
\]

\noindent For illustrative purposes the Markov chain is drawn in
the up-down direction in the diagram, with the horizontal direction
being used for the discrete time steps that are generated by the
MCMC update procedure. The $\pm n$ notation at the left hand side
shows the labelling convention that is used for the update that
occurs at each time step, where $+n$ indicates an interaction between
a node and its right hand neighbour (right is ``down'' in the diagram),
and $-n$ is the analogous notation for the left hand neighbour.
The $\pm n$ notation along the bottom of the diagram shows the actual
update interaction that occurs at each time step. The particular
sequence of MCMC updates that is represented in the diagram above
is unimportant because it is random.

There are 6 separate basic diagrams that are used to build the above
diagram which are shown in the diagram below. Usually a randomly
selected sequence of these diagrams forms the MCMC algorithm, but
other choices are possible.

\[
\begin{array}{ccccccccccccccccccccccccc}
 \cdot  & \longrightarrow  & \cdot  &   & \cdot  &   & \cdot  &
&   & \cdot  & \longrightarrow  & \cdot  &   & \cdot  & \longrightarrow
& \cdot  &   &   & \cdot  & \longrightarrow  & \cdot  &   & \cdot
& \longrightarrow  & \cdot  \\
   & \searrow  &   &   &   & \nearrow  &   &   &   &   &   &   &
&   &   &   &   &   &   & \ \  &   &   &   & \ \  &   \\
 \cdot  &   & \cdot  &   & \cdot  & \longrightarrow  & \cdot  &
&   & \cdot  & \longrightarrow  & \cdot  &   & \cdot  &   & \cdot
&   &   & \cdot  & \longrightarrow  & \cdot  &   & \cdot  & \longrightarrow
& \cdot  \\
   &   &   &   &   &   &   &   &   &   & \searrow  &   &   &   &
\nearrow  &   &   &   &   &   &   &   &   &   &   \\
 \cdot  & \longrightarrow  & \cdot  &   & \cdot  & \longrightarrow
& \cdot  &   &   & \cdot  &   & \cdot  &   & \cdot  & \longrightarrow
& \cdot  &   &   & \cdot  &   & \cdot  &   & \cdot  &   & \cdot
\\
   &   &   &   &   &   &   &   &   &   &   &   &   &   &   &   &
&   &   & \searrow  &   &   &   & \nearrow  &   \\
 \cdot  & \longrightarrow  & \cdot  &   & \cdot  & \longrightarrow
& \cdot  &   &   & \cdot  & \longrightarrow  & \cdot  &   & \cdot
& \longrightarrow  & \cdot  &   &   & \cdot  & \longrightarrow
& \cdot  &   & \cdot  & \longrightarrow  & \cdot  \\
   & +1 &   &   &   & -1 &   &   &   &   & +2 &   &   &   & -2 &
&   &   &   & +3 &   &   &   & -3 &
\end{array}
\]

The skeletonised structure of the diagrams can now be simplified
further to make it look more symmetrical as shown in the diagram
below, where the pieces of the above diagrams are drawn individually
in more symmetrical fashion.

\[
\begin{array}{ccccc}
 \begin{array}{ccc}
 \cdot  & \longrightarrow  & \cdot
\end{array} &   & \Longrightarrow  &   & \begin{array}{ccc}
 \longrightarrow  & \cdot  & \longrightarrow
\end{array} \\
   &   &   &   &   \\
   &   &   &   &   \\
 \begin{array}{ccc}
 \cdot  & \longrightarrow  & \cdot  \\
   & \searrow  &   \\
 \cdot  &   & \cdot
\end{array} &   & \Longrightarrow  &   & \begin{array}{ccc}
 \longrightarrow  & \cdot  & \longrightarrow  \\
   & \downarrow  &   \\
 \longrightarrow  & \cdot  & \longrightarrow
\end{array} \\
   &   &   &   &   \\
   &   &   &   &   \\
 \begin{array}{ccc}
 \cdot  & \longrightarrow  & \cdot  \\
   & \nearrow  &   \\
 \cdot  &   & \cdot
\end{array} &   & \Longrightarrow  &   & \begin{array}{ccc}
 \longrightarrow  & \cdot  & \longrightarrow  \\
   & \uparrow  &   \\
 \longrightarrow  & \cdot  & \longrightarrow
\end{array}
\end{array}
\]

\noindent This reduces the description of the MCMC algorithm to
a set of basic diagrams in which the state of a node evolves freely
(i.e. $\begin{array}{ccc}
 \longrightarrow  & \cdot  & \longrightarrow
\end{array}$) or is involved in an interaction (i.e. $\begin{array}{ccc}
 \longrightarrow  & \cdot  & \longrightarrow  \\
   & \downarrow  &
\end{array}$ and $\begin{array}{ccc}
 \longrightarrow  & \cdot  & \longrightarrow  \\
   & \uparrow  &
\end{array}$). These diagrams allow for the possibility that a node
has a ``memory'' of its previous state (i.e. an arrow comes in from
the left), so the MCMC diagrams above are a special case in which
this memory is discarded.

These diagrams can be used to represent higher order MCMC algorithms
which amalgamate the effect of several basic MCMC updates. Thus,
start by defining an MCMC update operator $\mathcal{H}$. For a \textit{pair}
of MRF nodes this is illustrated in Equation \ref{XRef-Equation-103125636},
which is of the form $\mathcal{H}\equiv \mathcal{I}+\mathcal{H}_{1}+\mathcal{H}_{2}$.
The $\mathcal{I}$ is the ``identity'' which corresponds to no update
occurring, and the $\mathcal{H}_{1}$ and $\mathcal{H}_{2}$ pieces
correspond to updates that occur on one or the other of the two
nodes, respectively.
\begin{equation}
\mathcal{H}\equiv \left( \begin{array}{ccc}
 \longrightarrow  & \cdot  & \longrightarrow  \\
   &   &   \\
 \longrightarrow  & \cdot  & \longrightarrow
\end{array}\right) +\left( \begin{array}{ccc}
 \longrightarrow  & \cdot  & \longrightarrow  \\
   & \downarrow  &   \\
 \longrightarrow  & \cdot  & \longrightarrow
\end{array}\right) +\left( \begin{array}{ccc}
 \longrightarrow  & \cdot  & \longrightarrow  \\
   & \uparrow  &   \\
 \longrightarrow  & \cdot  & \longrightarrow
\end{array}\right) %
\label{XRef-Equation-103125636}
\end{equation}

\noindent Multiple MC updates may then be generated by iterating
$\mathcal{H}$ to create powers of $\mathcal{H}$. For instance, $\mathcal{H}^{2}$
may be derived as by expanding out ${\{\mathcal{I}+\mathcal{H}_{1}+\mathcal{H}_{2}\}}^{2}$
and collecting together similar terms, as shown in Equation \ref{XRef-Equation-103125657}
and Equation \ref{XRef-Equation-1025104456}.
\begin{equation}
\mathcal{H}^{2}=A_{0}+A_{1}+A_{2}%
\label{XRef-Equation-103125657}
\end{equation}

\noindent where
\begin{equation}
\begin{array}{rl}
 A_{0} & \equiv \left( \begin{array}{ccccc}
 \longrightarrow  & \cdot  & \longrightarrow  & \cdot  & \longrightarrow
\\
   &   &   &   &   \\
 \longrightarrow  & \cdot  & \longrightarrow  & \cdot  & \longrightarrow
\end{array}\right)  \\
 A_{1} & \equiv \begin{array}{c}
 \left( \begin{array}{ccccc}
 \longrightarrow  & \cdot  & \longrightarrow  & \cdot  & \longrightarrow
\\
   & \downarrow  &   &   &   \\
 \longrightarrow  & \cdot  & \longrightarrow  & \cdot  & \longrightarrow
\end{array}\right) +\left( \begin{array}{ccccc}
 \longrightarrow  & \cdot  & \longrightarrow  & \cdot  & \longrightarrow
\\
   &   &   & \downarrow  &   \\
 \longrightarrow  & \cdot  & \longrightarrow  & \cdot  & \longrightarrow
\end{array}\right)  \\
 +\left( \begin{array}{ccccc}
 \longrightarrow  & \cdot  & \longrightarrow  & \cdot  & \longrightarrow
\\
   & \uparrow  &   &   &   \\
 \longrightarrow  & \cdot  & \longrightarrow  & \cdot  & \longrightarrow
\end{array}\right) +\left( \begin{array}{ccccc}
 \longrightarrow  & \cdot  & \longrightarrow  & \cdot  & \longrightarrow
\\
   &   &   & \uparrow  &   \\
 \longrightarrow  & \cdot  & \longrightarrow  & \cdot  & \longrightarrow
\end{array}\right)
\end{array} \\
 A_{2} & \equiv \begin{array}{c}
 \left( \begin{array}{ccccc}
 \longrightarrow  & \cdot  & \longrightarrow  & \cdot  & \longrightarrow
\\
   & \downarrow  &   & \downarrow  &   \\
 \longrightarrow  & \cdot  & \longrightarrow  & \cdot  & \longrightarrow
\end{array}\right) +\left( \begin{array}{ccccc}
 \longrightarrow  & \cdot  & \longrightarrow  & \cdot  & \longrightarrow
\\
   & \uparrow  &   & \uparrow  &   \\
 \longrightarrow  & \cdot  & \longrightarrow  & \cdot  & \longrightarrow
\end{array}\right)  \\
 +\left( \begin{array}{ccccc}
 \longrightarrow  & \cdot  & \longrightarrow  & \cdot  & \longrightarrow
\\
   & \downarrow  &   & \uparrow  &   \\
 \longrightarrow  & \cdot  & \longrightarrow  & \cdot  & \longrightarrow
\end{array}\right) +\left( \begin{array}{ccccc}
 \longrightarrow  & \cdot  & \longrightarrow  & \cdot  & \longrightarrow
\\
   & \uparrow  &   & \downarrow  &   \\
 \longrightarrow  & \cdot  & \longrightarrow  & \cdot  & \longrightarrow
\end{array}\right)
\end{array}
\end{array}%
\label{XRef-Equation-1025104456}
\end{equation}

\noindent The result in Equation \ref{XRef-Equation-103125657} and
Equation \ref{XRef-Equation-1025104456} may be simplified to Equation
\ref{XRef-Equation-913131538} and Equation \ref{XRef-Equation-1025104519}
(using $\mathcal{I}^{2}=\mathcal{I}$ and ${\mathcal{I}\mathcal{H}}_{i}=\mathcal{H}_{i}\mathcal{I}=\mathcal{H}_{i}$).
\begin{equation}
\mathcal{H}^{2}=B_{0}+B_{1}+A_{2}%
\label{XRef-Equation-913131538}
\end{equation}

\noindent where
\begin{equation}
\begin{array}{rl}
 B_{0} & \equiv \left( \begin{array}{ccc}
 \longrightarrow  & \cdot  & \longrightarrow  \\
   &   &   \\
 \longrightarrow  & \cdot  & \longrightarrow
\end{array}\right)  \\
 B_{1} & \equiv 2\left( \begin{array}{ccc}
 \longrightarrow  & \cdot  & \longrightarrow  \\
   & \downarrow  &   \\
 \longrightarrow  & \cdot  & \longrightarrow
\end{array}\right) +2\left( \begin{array}{ccc}
 \longrightarrow  & \cdot  & \longrightarrow  \\
   & \uparrow  &   \\
 \longrightarrow  & \cdot  & \longrightarrow
\end{array}\right)
\end{array}%
\label{XRef-Equation-1025104519}
\end{equation}

\noindent In the diagrammatic expression for $\mathcal{H}^{2}$ in
Equation \ref{XRef-Equation-1025104519} the first row represents
\textit{no} interaction, the second row \textit{one} interaction,
and the third row \textit{two} interactions. Note that the order
in which the interactions occur is important (i.e. $\mathcal{H}_{1}\mathcal{H}_{2}\neq
\mathcal{H}_{2}\mathcal{H}_{1}$ in general) so the diagrams in the
third row \textit{cannot} be combined. On the other hand ${\mathcal{I}\mathcal{H}}_{i}=\mathcal{H}_{i}\mathcal{I}=\mathcal{H}_{i}$
so the diagrams in the second row \textit{can} be combined.

These diagrams are actually Feynman diagrams, which describe operator
expressions in an visually appealing way. In this case they show
how the various operations invoked by the pieces of the MCMC update
operator $\mathcal{H}$ fit together in various ways to generate
the diagrammatic representation of the higher order MCMC update
operator $\mathcal{H}^{2}$. This example is simple enough that the
results are obvious, but the diagrammatic technique generalises
to arbitrarily complicated cases.
\section{Applications of the MCMC Update Operator}\label{XRef-Section-11423193}

The aim of this section is to show some simple practical uses of
the operator approach that is described in Section \ref{XRef-Subsection-33113632}.
No attempt will be made to do extensive computations, because these
will be presented in future papers in this ``discrete network dynamics''
series of papers.

Section \ref{XRef-Subsubsection-41111651} illustrates how the MCMC
update operator correctly generates MCMC updates for histograms
that are each occupied by a single sample, thus ensuring backwards
compatibility between the operator approach and the standard MCMC
algorithm for sampling MRFs. Section \ref{XRef-Subsubsection-41115422}
generalises this to the case of multiply occupied states, and derives
the equilibrium state of a single node MRF which has the same properties
as ACEnet \cite{Luttrell1996}.
\subsection{Update of Single-Sample States}\label{XRef-Subsubsection-41111651}

As a check on the result for $\mathcal{H}$ in Equation \ref{XRef-Equation-41104529}
verify that the application of $\mathcal{H}$ to a standard MRF state
(i.e. {\itshape one} sample per node) leads to the expected standard
form of the MCMC update.

In a standard MRF only a single bin $i_{u}$ is occupied at each
node $u$. For an $N$-node MRF this defines a {\itshape pure} state
$\Psi ( i_{1},i_{2},\cdots ,i_{N}) $ that has the form
\begin{equation}
\Psi ( i_{1},i_{2},\cdots ,i_{N}) \equiv \left( \prod _{u=1}^{N}a_{i_{u}}^{u
\dagger }\right) \left. \left| 0\right. \right\rangle
\end{equation}

\noindent The first operator to consider in Equation \ref{XRef-Equation-41104529}
is the number operator $\mathcal{N}_{k}^{t}$ (for measuring how
many samples are in bin $k$ at node $t$). When $\mathcal{N}_{k}^{t}$
is applied to $\Psi ( i_{1},i_{2},\cdots ,i_{N}) $ it gives
\begin{equation}
\begin{array}{rl}
 \mathcal{N}_{k}^{t} \Psi ( i_{1},i_{2},\cdots ,i_{N})  & =\mathcal{N}_{k}^{t}
\left( \prod _{u=1}^{N}a_{i_{u}}^{u \dagger }\right) \left. \left|
0\right. \right\rangle   \\
  & =\delta _{i_{t},k}\Psi ( i_{1},i_{2},\cdots ,i_{N})
\end{array}
\end{equation}

\noindent so the number $\delta _{i_{t},k}$ is 1 if the bin at node
$t$ being examined (i.e. $k$) matches the bin in which the sample
at node $t$ is to be found (i.e. $i_{t}$), and is 0 otherwise.

Insert this result into the $\prod _{t\in C( s) }\sum _{k=1}^{m_{t}}p_{i,k}^{s,t}\mathcal{N}_{k}^{t}$
part of $\mathcal{H}$ in Equation \ref{XRef-Equation-41104529} to
obtain the following simplification
\begin{equation}
\begin{array}{rl}
 \left( \prod _{t\in C( s) }\sum _{k=1}^{m_{t}}p_{i,k}^{s,t}\mathcal{N}_{k}^{t}\right)
\Psi ( i_{1},i_{2},\cdots ,i_{N})  & =\left( \prod _{t\in C( s)
}\sum _{k=1}^{m_{t}}p_{i,k}^{s,t}\mathcal{N}_{k}^{t}\right)  \left(
\prod _{u=1}^{N}a_{i_{u}}^{u \dagger }\right) \left. \left| 0\right.
\right\rangle   \\
  & =\left( \prod _{t\in C( s) }\sum _{k=1}^{m_{t}}p_{i,k}^{s,t}\delta
_{i_{t},k}\right) \Psi ( i_{1},i_{2},\cdots ,i_{N})  \\
  & =\left( \prod _{t\in C( s) }p_{i,i_{t}}^{s,t}\right) \Psi (
i_{1},i_{2},\cdots ,i_{N})
\end{array}%
\label{XRef-Equation-4111533}
\end{equation}

\noindent which is equal to $\Psi ( i_{1},i_{2},\cdots ,i_{N}) $
weighted by the product of the 2-clique factors that involve node
$s$. This result correctly computes the 2-clique influence of the
neighbours of node $s$ that is expected in a standard MCMC algorithm.

$\mathcal{H}$ in Equation \ref{XRef-Equation-41104529} also involves
the transition operator $\mathcal{T}_{i,j}^{s}$. Apply $\mathcal{T}_{i,j}^{s}$
to $\Psi ( i_{1},i_{2},\cdots ,i_{N}) $ to obtain
\begin{equation}
\begin{array}{rl}
 \mathcal{T}_{i,j}^{s} \Psi ( i_{1},i_{2},\cdots ,i_{N})  & =\mathcal{T}_{i,j}^{s}
\left( \prod _{u=1}^{N}a_{i_{u}}^{u \dagger }\right) \left. \left|
0\right. \right\rangle   \\
  & =a_{i}^{s \dagger }a_{j}^{s} a_{i_{1}}^{1 \dagger }a_{i_{2}}^{2
\dagger }\cdots  a_{i_{s}}^{s \dagger } \cdots \ \ a_{i_{N}}^{N
\dagger }\left. \left| 0\right. \right\rangle   \\
  & =a_{i}^{s \dagger }a_{i_{1}}^{1 \dagger }a_{i_{2}}^{2 \dagger
}\cdots  \left( a_{i_{s}}^{s \dagger }a_{j}^{s}+\delta _{i_{s},j}\right)
\cdots \ \ a_{i_{N}}^{N \dagger } \left. \left| 0\right. \right\rangle
\\
  & =\begin{array}{c}
 a_{i}^{s \dagger }a_{i_{1}}^{1 \dagger }a_{i_{2}}^{2 \dagger }\cdots
a_{i_{s}}^{s \dagger }\cdots \ \ a_{i_{N}}^{N \dagger }a_{j}^{s}
\left. \left| 0\right. \right\rangle   \\
 +\delta _{i_{s},j}a_{i_{1}}^{1 \dagger }a_{i_{2}}^{2 \dagger }\cdots
a_{i}^{s \dagger } \cdots \ \ a_{i_{N}}^{N \dagger } \left. \left|
0\right. \right\rangle
\end{array} \\
  & =\delta _{i_{s},j}\Psi ( i_{1},i_{2},\cdots ,i_{s-1},i,i_{s+1},\cdots
,i_{N})
\end{array}%
\label{XRef-Equation-4111546}
\end{equation}

\noindent where the annihilation operator $a_{j}^{s}$ is moved to
the right, picking up a non-zero commutator only when it moves past
the creation operator $a_{i_{s}}^{s \dagger }$ (i.e. both the creation
and the annihilation are at the {\itshape same} node so they do
{\itshape not} commute if $i_{s}=j$), and finally meets the empty
state $|0\rangle $ which it annihilates. This result is equal to
$\Psi ( i_{1},i_{2},\cdots ,i_{s-1},i,i_{s+1},\cdots ,i_{N}) $ weighted
by a factor $\delta _{i_{s},j}$, which corresponds to a new pure
state in which the sample at node $s$ has hopped to bin $i$, weighted
by 1 if the sample at node $s$ started off in bin $j$, and 0 otherwise.
This is exactly the behaviour that is expected of the transition
operator $\mathcal{T}_{i,j}^{s}$.

Finally, inserting the results in Equation \ref{XRef-Equation-4111533}
and Equation \ref{XRef-Equation-4111546} into $\mathcal{H}$ in Equation
\ref{XRef-Equation-41104529} gives
\begin{equation}
\begin{array}{rl}
 \mathcal{H} \Psi ( i_{1},i_{2},\cdots ,i_{N})  & =\sum _{s=1}^{N}\sum
_{i,j=1}^{m_{s}}\mathcal{T}_{i,j}^{s} \left( \prod _{t\in C( s)
}\sum _{k=1}^{m_{t}}p_{i,k}^{s,t}\mathcal{N}_{k}^{t}\right)  \Psi
( i_{1},i_{2},\cdots ,i_{N})  \\
  & =\sum _{s=1}^{N}\sum _{i,j=1}^{m_{s}}\delta _{i_{s},j} \left(
\prod _{t\in C( s) }p_{i,i_{t}}^{s,t}\right)  \Psi ( i_{1},i_{2},\cdots
,i_{s-1},i,i_{s+1},\cdots ,i_{N})  \\
  & =\sum _{s=1}^{N}\sum _{i=1}^{m_{s}}\left( \prod _{t\in C( s)
}p_{i,i_{t}}^{s,t}\right)  \Psi ( i_{1},i_{2},\cdots ,i_{s-1},i,i_{s+1},\cdots
,i_{N})
\end{array}%
\label{XRef-Equation-41111742}
\end{equation}

\noindent The action of $\mathcal{H}$ on the \textit{pure} state
$\Psi ( i_{1},i_{2},\cdots ,i_{N}) $ produces a weighted sum of
states (or \textit{mixed} state), because the effect of $\mathcal{H}$
at each node $s$ is to {\itshape simultaneously} create $m_{s}$
states $\Psi ( i_{1},i_{2},\cdots ,i_{s-1},i,i_{s+1},\cdots ,i_{N})
$ (for $i=1,2,\cdots ,m_{s}$), each of which has its own probability
factor $\prod _{t\in C( s) }p_{i,i_{t}}^{s,t}$ (i.e. product of
2-clique factors), which is a total of $m_{1}m_{2}\cdots  m_{N}$
states with their corresponding probability factors. Note that this
ensemble of histograms should be regrouped so that multiple copies
of identical histograms are represented as a single copy with an
appropriate weighting factor. Thus $\mathcal{H} \Psi ( i_{1},i_{2},\cdots
,i_{N}) $ is {\itshape precisely} the ensemble of states from which
the standard MCMC update algorithm draws its updated state.

This verifies that the update operator $\mathcal{H}$ generates the
correct behaviour when only a \textit{single} bin $i_{u}$ is occupied
at each node $u$, as is the case in standard MCMC simulations of
MRFs. Similarly, higher order cliques produce the same consistency
between what the update operator $\mathcal{H}$ generates and what
the standard MCMC algorithm generates, so the assumed operator form
of $\mathcal{H}$ is backwardly compatible with MCMC simulations
of standard MRFs with a single sample per node.

Standard MCMC algorithms randomly select a \textit{single} state
from the above ensemble of states generated by the action of the
update operator $\mathcal{H}$; the probability of a particular state
being selected is given by the probability factor that weights that
state in the ensemble. More sophisticated MCMC algorithms, known
as particle filtering algorithms \cite{DoucetGodsillAndrieu2000},
select \textit{several} states from the ensemble which allows several
alternative updates to be simultaneously followed, which allows
the probability over alternatives to be represented in a sampled
form. However, all of these approaches fit into the same theoretical
framework where the update operator $\mathcal{H}$ generates the
\textit{full} ensemble of alternatives.

Note that pure states and mixed states are related to doubly distributional
population codes \cite{SahaniDayan2003}. Thus a pure state specifies
a single joint state of the MRF nodes, whereas a mixed state specifies
a range of alternative joint states of the MRF nodes. The operator
algebra presented in this paper provides a complete and consistent
framework for using MCMC algorithms to manipulate these pure and
mixed MRF states, or equivalently the corresponding doubly distributional
population codes.
\subsection{Equilibrium Multi-Sample State}\label{XRef-Subsubsection-41115422}

The aim of this section is to demonstrate in detail that the MCMC
update operator $\mathcal{H}\equiv \sum _{i=1}^{m}p_{i}{a_{i}}^{\dagger
}\sum _{j=1}^{m}a_{j}$ has an equilibrium state which has the same
properties as ACEnet \cite{Luttrell1996}.

In Section \ref{XRef-Subsubsection-41111651} the application of
$\mathcal{H}$ to a pure state $\Psi ( i_{1},i_{2},\cdots ,i_{N})
$ converts it into a mixed state (see Equation \ref{XRef-Equation-41111742}).
The aim now is to derive the equilibrium mixed state that self-consistently
maps to itself under the action of $\mathcal{H}$. This would correspond
to a mixed state that contains exactly the right mixture of pure
states to balance the hopping rates generated by $\mathcal{H}$.
In physics this is known as the {\itshape detailed balance} condition.
When there is a {\itshape single} sample per node this equilibrium
mixed state corresponds to the equilibrium ensemble that the standard
MCMC update algorithm seeks to generate.

It is {\itshape not} possible in general to analytically derive
this equilibrium mixed state; if it were then MCMC algorithms would
not be needed. This intractability arises because the clique factors
cause the samples at neighbouring nodes (i.e. nodes in the same
clique) to interact with each other, which leads to the development
of {\itshape indirect} long-range correlations between nodes by
cascading together multiple {\itshape direct} short-range interactions
(i.e. paths of influence are built out of interlinked clique factors).
The summation over all possible paths via which the nodes can interact
indirectly with each other is {\itshape not} analytically tractable,
except in simple cases such as when the nodes interact along a 1-dimensional
chain (or any acyclic graph of interactions). More interesting cases,
such as 2-dimensional sheets of node interactions, are {\itshape
not} analytically tractable in general (although there are special
cases that are exceptions, such as the 2-dimensional Ising model).

One case which {\itshape can} be solved analytically is the case
of an MRF with a \textit{single} node that interacts with a fixed
external source. In effect, this is an $N$-node MRF in which $N-1$
of the nodes are frozen, and their influence on the single remaining
(unfrozen) node is represented by the external source. This case
is interesting because it is the model that is used in the simplest
version (i.e. single coding layer) of ACEnet \cite{Luttrell1996};
it is therefore prudent to use the operator methods developed in
this paper to verify that the MCMC equilibrium state corresponds
to the behaviour that is observed in ACEnet.\,

The state space of a multiply occupied 1-node MRF is an $n$-sample
histogram. The aim now is to derive the equilibrium state of an
$n$-sample histogram under the action of repeated MCMC samplings
generated by $\mathcal{H}=\sum _{i=1}^{m}p_{i}{a_{i}}^{\dagger }\sum
_{j=1}^{m}a_{j}$ (see Equation \ref{XRef-Equation-1018122141}),
where the probabilities $p_{i}$ are derived from a fixed external
source. The equilibrium mixed state $\Psi $ must satisfy the self-consistent
bound state equation
\begin{equation}
\left( \sum _{j=1}^{m}p_{j}{a_{j}}^{\dagger }\right) \left( \sum
_{i=1}^{m}a_{i}\right)  \Psi =\lambda  \Psi %
\label{XRef-Equation-323111626}
\end{equation}

\noindent where $\lambda $ is an eigenvalue. In other words the
MCMC update operator must map the equilibrium state into a multiple
of itself, as is expected of an equilibrium state. Because correct
normalisation of the state and of the MCMC update operator have
not been imposed (to avoid lots of distracting normalisation factors
appearing in the mathematics), the eigenvalue is not the expected
$\lambda =1$, but nevertheless the value of $\lambda $ may be readily
interpreted (see after Equation \ref{XRef-Equation-1021103352}).

The mixed state $\Psi $ can be expanded as a weighted mixture of
pure states thus
\begin{equation}
\Psi =\sum _{n_{1},n_{2},\cdots ,n_{m}}\psi ( n_{1},n_{2},\cdots
,n_{m}) \prod _{k=1}^{m}{\left( {a_{k}}^{\dagger }\right) }^{n_{k}}\left.
\left| 0\right. \right\rangle  %
\label{XRef-Equation-323114558}
\end{equation}

\noindent where ${({a_{k}}^{\dagger })}^{n_{k}}|0\rangle $ is (up
to a normalising constant) a histogram with $n_{k}$ samples in bin
$k$, $\prod _{k=1}^{m}{({a_{k}}^{\dagger })}^{n_{k}}|0\rangle $
is (up to a normalising constant) a histogram with occupancy $(n_{1},n_{2},\cdots
,n_{m})$, $\psi ( n_{1},n_{2},\cdots ,n_{m}) $ is the probability
(up to a normalising constant) of this histogram occurring, and
$\sum _{n_{1},n_{2},\cdots ,n_{m}}(\cdots )$ is a mixture of such
histograms. Note that it is not necessary to introduce the normalising
constants explicitly because all we are trying to do is to demonstrate
that $\Psi $ is a solution of Equation \ref{XRef-Equation-323111626}.

First of all, force the {\itshape total} number of samples to be
constrained. In physicists' terminology, the case with a fixed number
of samples is a {\itshape canonical} ensemble, rather than a {\itshape
grand canonical} ensemble in which the total number of samples would
be allowed to vary. Thus write $\Psi $ as
\begin{equation}
\Psi =\sum _{n_{1},n_{2},\cdots ,n_{m}}\delta _{n,n_{1}+n_{2}+ \cdots
+n_{m}}\psi ( n_{1},n_{2},\cdots ,n_{m}) \prod _{k=1}^{m}{\left(
{a_{k}}^{\dagger }\right) }^{n_{k}}\left. \left| 0\right. \right\rangle
\label{XRef-Equation-41115157}
\end{equation}

\noindent where the Kronecker delta $\delta _{n,n_{1}+n_{2}+ \cdots
+n_{m}}$ ensures that only terms in $\sum _{n_{1},n_{2},\cdots ,n_{m}}(\cdots
)$ that satisfy the condition $n=n_{1}+n_{2}+ \cdots  +n_{m}$ can
contribute.

Now find the state $\Psi $ that satisfies the consistency condition
in Equation \ref{XRef-Equation-323111626}. First substitute Equation
\ref{XRef-Equation-41115157} into the left hand side of Equation
\ref{XRef-Equation-323111626} to obtain
\begin{equation}
\sum _{n_{1},n_{2},\cdots ,n_{m}}\delta _{n,n_{1}+n_{2}+ \cdots
+n_{m}}\psi ( n_{1},n_{2},\cdots ,n_{m})  \left( \sum _{j=1}^{m}p_{j}{a_{j}}^{\dagger
}\right) \left( \sum _{i=1}^{m}a_{i}\right) \prod _{k=1}^{m}{\left(
{a_{k}}^{\dagger }\right) }^{n_{k}}\left. \left| 0\right. \right\rangle
\label{XRef-Equation-41115256}
\end{equation}

\noindent Now use that $a_{i} {({a_{j}}^{\dagger })}^{n}|0\rangle
=n \delta _{i,j} {({a_{j}}^{\dagger })}^{n-1}|0\rangle $ to move
all of the annihilation operators to the right in the $(\sum _{j=1}^{m}p_{j}{a_{j}}^{\dagger
})(\sum _{i=1}^{m}a_{i})\prod _{k=1}^{m}{({a_{k}}^{\dagger })}^{n_{k}}|0\rangle
$ part of the expression in Equation \ref{XRef-Equation-41115256}
to obtain the following simplification
\begin{equation}
\begin{array}{rl}
 \left. \left( \cdots \right) \left| 0\right. \right\rangle   &
=\left( \sum _{j=1}^{m}p_{j}{a_{j}}^{\dagger }\right) \sum _{i=1}^{m}n_{i}
{\left( {a_{1}}^{\dagger }\right) }^{n_{1}} \cdots  {\left( {a_{i}}^{\dagger
}\right) }^{n_{i}-1} \cdots  {\left( {a_{m}}^{\dagger }\right) }^{n_{m}}\left.
\left| 0\right. \right\rangle   \\
  & =\sum _{j=1}^{m}p_{j} \left( \begin{array}{c}
 n_{j} {\left( {a_{1}}^{\dagger }\right) }^{n_{1}} \cdots  {\left(
{a_{m}}^{\dagger }\right) }^{n_{m}}\left. \left| 0\right. \right\rangle
\\
 +\sum _{\begin{array}{l}
 i=1, \\
 i\neq j
\end{array}}^{m}n_{i} {\left( {a_{1}}^{\dagger }\right) }^{n_{1}}
\cdots  {\left( {a_{i}}^{\dagger }\right) }^{n_{i}-1} \cdots  {\left(
{a_{j}}^{\dagger }\right) }^{n_{j}+1} \cdots  {\left( {a_{m}}^{\dagger
}\right) }^{n_{m}}\left. \left| 0\right. \right\rangle
\end{array}\right)
\end{array}
\end{equation}

\noindent where the cases $i=j$ (annihilation and creation within
a single bin) and $i\neq j$ (annihilation in one bin and creation
in another bin, i.e. hopping) have to be considered separately.

The contribution for a given final state $j$ (but summing over the
initial state $i$) can be represented diagrammatically as follows
\[
\begin{array}{ccccc}
 p_{j}n_{j} & \left( \begin{array}{ccccc}
 i \left( =j\right)  & \overset{a_{i}}{\longrightarrow } & \cdot
& \overset{{a_{j}}^{\dagger }}{\longrightarrow } & j \\
   &   & \Uparrow  &   &   \\
   &   & \mathrm{source} &   &
\end{array}\right)  & + & p_{j}\sum _{\begin{array}{l}
 i=1, \\
 i\neq j
\end{array}}^{m}n_{i} & \left( \begin{array}{ccccc}
 i \left( \neq j\right)  &   &   &   &   \\
   & \overset{a_{i}}{\searrow } &   &   &   \\
   &   & \cdot  & \overset{{a_{j}}^{\dagger }}{\longrightarrow }
& j \\
   &   & \Uparrow  &   &   \\
   &   & \mathrm{source} &   &
\end{array}\right)
\end{array}
\]

\noindent which is a sum of contributions of the form
\[
p_{j}\begin{array}{cc}
 n_{i} & \left( \begin{array}{ccccc}
 i &   &   &   &   \\
   & \overset{a_{i}}{\searrow } &   &   &   \\
   &   & \cdot  & \overset{{a_{j}}^{\dagger }}{\longrightarrow }
& j \\
   &   & \Uparrow  &   &   \\
   &   & \mathrm{source} &   &
\end{array}\right)
\end{array}
\]

\noindent where the overall factor of $n_{i}$ comes from the fact
that the annihilation operator $a_{i}$ has $n_{i}$ samples to choose
from in the initial state.

The coefficients of corresponding contributions to the left hand
side and right hand side of the equilibrium condition in Equation
\ref{XRef-Equation-323111626} can now be matched up. Note that this
matching of coefficients is allowed because the set of states $\prod
_{k=1}^{m}{({a_{k}}^{\dagger })}^{n_{k}}|0\rangle $ is orthogonal
and complete (see Section \ref{XRef-Subsubsection-11714376}). This
leads to the following consistency equation that interrelates the
$\psi ( n_{1},n_{2},\cdots ,n_{m}) $.
\begin{multline}
\sum _{j=1}^{m}p_{j} \left( \begin{array}{c}
 n_{j} \psi ( n_{1},n_{2},\cdots ,n_{m})  \\
 +\sum _{\begin{array}{l}
 i=1, \\
 i\neq j
\end{array}}^{m}\left( n_{i}+1\right)  \psi ( n_{1},\cdots  ,n_{i}+1,\cdots
,n_{j}-1,\cdots ,n_{m})
\end{array}\right) \\
=\lambda  \psi ( n_{1},n_{2},\cdots ,n_{m}) %
\label{XRef-Equation-41132542}
\end{multline}

\noindent Now define a trial solution to this equation (where $n=n_{1}+n_{2}+
\cdots  +n_{m}$)
\begin{equation}
\psi ( n_{1},n_{2},\cdots ,n_{m}) =\frac{n!}{n_{1}!n_{2}!\cdots
n_{m}!}{p_{1}}^{n_{1}}{p_{2}}^{n_{2}}\cdots  {p_{m}}^{n_{m}}%
\label{XRef-Equation-323114626}
\end{equation}

\noindent This trial solution corresponds to placing $n$ samples
at random into the histogram, using sampling probabilities $(p_{1},p_{2},\cdots
,p_{m})$ for each of the $m$ bins. The probability factor ${p_{1}}^{n_{1}}{p_{2}}^{n_{2}}\cdots
{p_{m}}^{n_{m}}$ is the probability of each possible way of placing
$n$ samples (taking account of the order in which the samples are
placed), and the multinomial factor $\frac{n!}{n_{1}!n_{2}!\cdots
n_{m}!}$ is the number of possible orderings of samples that leave
the histogram unchanged (i.e. permute \textit{within} bins but not
\textit{between} bins). It is reasonable to expect this to be the
solution because the effect of $\mathcal{H}$ (i.e. $\sum _{i=1}^{m}p_{i}{a_{i}}^{\dagger
}\sum _{j=1}^{m}a_{j}$) is to randomly annihilate a sample from
the histogram, and then to create it again with probability $p_{i}$
in bin $i$ (which is a \textit{memoryless} operation), so the $\psi
( n_{1},n_{2},\cdots ,n_{m}) $ given in Equation \ref{XRef-Equation-323114626}
should be an equilibrium solution for updates generated by $\mathcal{H}$.

Substitute this trial solution into the consistency equation Equation
\ref{XRef-Equation-41132542} to obtain
\begin{multline}
\sum _{j=1}^{m}p_{j} \left( \begin{array}{c}
 n_{j} \frac{n!}{n_{1}!\cdots  n_{m}!}{p_{1}}^{n_{1}}\cdots  {p_{m}}^{n_{m}}
\\
 +\sum _{\begin{array}{l}
 i=1, \\
 i\neq j
\end{array}}^{m}\left( n_{i}+1\right)  \frac{n!}{n_{1}!\cdots  \left(
n_{i}+1\right) !\cdots  \left( n_{j}-1\right) !\cdots  n_{m}!}{p_{1}}^{n_{1}}\cdots
{p_{i}}^{n_{i}+1}\cdots  {p_{j}}^{n_{j}-1}\cdots  {p_{m}}^{n_{m}}
\end{array}\right) \\
=\lambda  \frac{n!}{n_{1}!\cdots  n_{m}!}{p_{1}}^{n_{1}}\cdots
{p_{m}}^{n_{m}}
\end{multline}

\noindent Cancel the factorials and the probability factors.
\begin{equation}
\sum _{j=1}^{m}p_{j} \left( n_{j}+\sum _{\begin{array}{l}
 i=1, \\
 i\neq j
\end{array}}^{m}n_{j}\frac{p_{i}}{p_{j}}\right) =\lambda
\end{equation}

\noindent Solve this equation for the eigenvalue $\lambda $, and
use that $\sum _{i=1}^{m} p_{i}=1$ and $\sum _{j=1}^{m}n_{j}=n$
to simplify the result.
\begin{equation}
\begin{array}{rl}
 \lambda  & =\sum _{j=1}^{m}p_{j} n_{j}+\sum _{j=1}^{m} \sum _{\begin{array}{l}
 i=1, \\
 i\neq j
\end{array}}^{m}p_{i}n_{j} \\
  & =\sum _{j=1}^{m}p_{j} n_{j}+\left( \sum _{i,j=1}^{m} p_{i}n_{j}-\sum
_{j=1}^{m}p_{j}n_{j}\right)  \\
  & =\left( \sum _{i=1}^{m} p_{i}\right)  \left( \sum _{j=1}^{m}n_{j}\right)
\\
  & =n
\end{array}%
\label{XRef-Equation-1021103352}
\end{equation}

\noindent Thus $\lambda =n$ which is the (fixed) total number of
samples in the histogram. The source of this factor is $\mathcal{H}\equiv
\sum _{i=1}^{m}p_{i}{a_{i}}^{\dagger }\sum _{j=1}^{m}a_{j}$, where
each annihilation operator $a_{j}$ has $n_{j}$ to choose from in
the initial state, so the sum of annihilation operators $\sum _{j=1}^{m}a_{j}$
generates $\sum _{j=1}^{m}n_{j}=n$ separate contributions. The fact
that $\lambda $ is a constant means that the consistency equation
(i.e. Equation \ref{XRef-Equation-41132542}) has an eigenvalue $\lambda
$ that is \textit{independent} of the choice of $(n_{1},n_{2},\cdots
,n_{m})$, which means that the update operator $\mathcal{H}$ has
the \textit{same} effect on each pure state component of the equilibrium
state $\Psi $ (as is required in order for $\Psi $ to satisfy Equation
\ref{XRef-Equation-323111626}).

The result in Equation \ref{XRef-Equation-1021103352} verifies that
the trial solution proposed in Equation \ref{XRef-Equation-323114626}
is correct, and that the equilibrium histogram state corresponds
to placing $n$ samples at random into the histogram using sampling
probabilities $(p_{1},p_{2},\cdots ,p_{m})$ for each of the $m$
bins.

Summarise these results:
\begin{enumerate}
\item Basic MCMC update operator: $\mathcal{H}=(\sum _{j=1}^{m}p_{j}{a_{j}}^{\dagger
})(\sum _{i=1}^{m}a_{i})$
\item General state (fixed $n$): $\Psi =\sum _{n_{1},n_{2},\cdots
,n_{m}}\delta _{n,n_{1}+n_{2}+ \cdots  +n_{m}}\psi ( n_{1},n_{2},\cdots
,n_{m}) \prod _{k=1}^{m}{({a_{k}}^{\dagger })}^{n_{k}}|0\rangle
$
\item Equilibrium condition: $(\sum _{j=1}^{m}p_{j}{a_{j}}^{\dagger
})(\sum _{i=1}^{m}a_{i}) \Psi =\lambda  \Psi $
\item Equilibrium state: $\psi ( n_{1},n_{2},\cdots ,n_{m}) =\frac{n!}{n_{1}!n_{2}!\cdots
n_{m}!}{p_{1}}^{n_{1}}{p_{2}}^{n_{2}}\cdots  {p_{m}}^{n_{m}}$ with
$\lambda =n$
\end{enumerate}

The equilibrium state is a {\itshape mixture} of pure states, where
each pure state is weighted by the probability of its occurrence.
In this approach the state $\Psi $ of the system corresponds to
the {\itshape entire} probability-weighted ensemble of alternative
histograms. In effect, these histograms mix with each other under
the updating action of the fixed external source that causes the
samples in the bins of each histogram to hop from bin to bin, whilst
conserving the total number of samples in the histogram (i.e. there
is migration of samples but no birth or death of samples). The equilibrium
condition ensures that the mixing that occurs due to the hopping
of samples has no net effect on the probability-weighted ensemble
of alternative histograms.

This completes the demonstration that the simplest (i.e. a single
node) multiple occupancy MRF has the same properties as ACEnet \cite{Luttrell1996},
which is \textit{defined} as having an equilibrium state that is
generated by the random (but probability-weighted) placement of
$n$ samples into a set of histogram bins. Also, larger SONs can
be built out of multiple linked ACEnet modules, and these correspond
to MRFs with a larger number of nodes. This unification of MRFs
and SONs is possible because both approaches can be viewed as implementing
algorithms for manipulating samples in histogram bins, and all such
algorithms can be expressed by using the algebra of creation and
annihilation operators. A key advantage of this MRF/SON unification
is that the techniques that are used to train SONs (i.e. to discover
structure in data) can now be used to train MRFs, which allows the
MRF graph structure (i.e. nodes and connections) to adapt itself
so that it is better matched to the data it is trying to model.

The MCMC updating of MRFs whose nodes are occupied by multiple samples
potentially leads to lots of interesting properties. The derivation
above shows how a single node MRF behaves under the influence of
a \textit{fixed} external source, but more interesting behaviour
occurs when either the MRF has a single node but the external source
is \textit{variable}, or if the MRF has multiple interacting nodes
so that each node sees the variable state of the other nodes. This
last case is especially interesting in MRFs that are trained as
SONs, because it leads to behaviours in which the samples that occupy
the nodes act collectively, and thus cause the joint node states
to behave like extended symbols (see Section \ref{XRef-Subsubsection-331134331}
for some diagrams that illustrate this point in more detail).
\section{Conclusions}

The work described in this paper assumes that Markov random field
models are used to implement Bayesian inference. The key contribution
of this paper is an implementation using creation and annihilation
operators of MCMC algorithms for simulating MRFs. This theoretical
framework has a similar structure to that used in quantum field
theories of bosons in physics \cite{Zee2003}. An equilibrium solution
of the MCMC update operator is derived which is shown to be equivalent
to the equilibrium behaviour of the adaptive cluster expansion network
(ACEnet) \cite{Luttrell1996}, which is a type of self-organising
network that computes using discrete-valued quantities.

This point of contact between MRF theory and SON behaviour allows
the theories of these two fields to be unified. Although MRFs and
SONs are superficially \textit{different} (MRFs have one sample
per node, whereas ACEnet SONs have multiple samples per node), the
underlying operators that are used to manipulate them are the \textit{same}.
MRF theory could benefit from this unification by being able to
make use of SONs to build MRF networks in a data-driven way. SON
theory could benefit from this unification by being able to make
full use of the rich theoretical theory of MRFs.

It is very convenient that MRFs and SONs are unified within a QFT
framework, because such theories are used extensively by physicists
to describe the interaction of particles, and many techniques have
been developed to compute results using such theories. We have found
that it is very easy to transfer knowledge from QFT to the unified
MRF/SON framework presented in this paper. Also, the diagrammatic
notation (i.e. Feynman diagrams) makes it much easier to understand
what MCMC algorithms are actually doing, without becoming submerged
in large amounts of theory.

Future papers in this ``discrete network dynamics'' series of papers
will focus in detail on the consequences of implementing MCMC algorithms
using update operators built out of creation and annihilation operators.
\section{Acknowledgements}

This research presented in this paper was supported by the United
Kingdom's MoD Corporate Research Programme.

\appendix

\end{document}